\documentclass[journal,twoside,web]{ieeecolor}
\usepackage{tmi}
\usepackage{cite}
\usepackage{subfigure}
\usepackage{amsmath,amssymb,amsfonts}
\usepackage{chemarrow}
\usepackage{graphicx}
\usepackage{textcomp}
\usepackage{colortbl}
\usepackage[lined,ruled,vlined]{algorithm2e}
\definecolor{mycommentcolor}{RGB}{191, 191, 191}

\SetCommentSty{mycomment}
\SetKwComment{Comment}{/* }{ */}
\usepackage[colorlinks=true,
    linkcolor=blue,
urlcolor=blue,citecolor=blue]{hyperref}
\usepackage{longtable}
\newlength\savewidth\newcommand\shline{\noalign{\global\savewidth\arrayrulewidth
  \global\arrayrulewidth 1.5pt}\hline\noalign{\global\arrayrulewidth\savewidth}}
\usepackage{booktabs}
\usepackage{bbding}
\usepackage{latexsym}
\usepackage{framed,multirow}
\usepackage{threeparttable}
\usepackage{xspace}
\usepackage{cite}
\usepackage{mathtools}
\usepackage{chemarrow}

\usepackage{xargs} 
\usepackage[pdftex,dvipsnames]{xcolor}

\usepackage{stmaryrd}
\newcommand{\conc}[2]{{\left\llbracket#1, #2\right\rrbracket}}

\usepackage{algpseudocode}
\usepackage{bm}
\usepackage{bbm}

\newcommand{\ie}{\textit{i}.\textit{e}.}
\newcommand{\eg}{\textit{e}.\textit{g}.}

\newcommand{\eat}[1]{}

\newcommand{\tabref}[1]{Table~\ref{#1}}
\newcommand{\figref}[1]{Fig.~\ref{#1}}

\newcommand{\equref}[1]{Eq.~\eqref{#1}}
\newcommand{\secref}[1]{Sec.~\ref{#1}}

\newcommand{\loss}[1]{\mathcal{L}_\text{#1}}

\newcommand{\lamda}[1]{\lambda_\text{#1}}

\newcommand{\best}[1]{{\textbf{#1}}}
\newcommand{\suboptimal}[1]{{{#1}}}

\newcommand{\Set}[1]{\mathcal{#1}}
\newcommand{\vct}[1]{{\mathbf #1}}
\newcommand{\mat}[1]{{\mathbf #1}}
\newcommand{\img}[1]{{\mathbf #1}}

\newcommand{\methodname}{{HiDiff}\xspace}
\newcommand{\refiner}{{BBDM}\xspace}
\newcommand{\methodbase}{{BerDiff}\xspace}

\definecolor{tablecolor}{rgb}{0.906, 0.89, 1}
\definecolor{tablecolor_1}{rgb}{0.878, 0.933, 0.933}

\def\BibTeX{{\rm B\kern-.05em{\sc i\kern-.025em b}\kern-.08em
    T\kern-.1667em\lower.7ex\hbox{E}\kern-.125emX}}
\begin{document}
\title{\methodname: Hybrid Diffusion Framework for \\Medical Image Segmentation}

\author{
Tao Chen, 
Chenhui Wang,
Zhihao Chen,
Yiming Lei,~\IEEEmembership{Member, IEEE},\\\
Hongming Shan,~\IEEEmembership{Senior Member, IEEE}
\thanks{This work was supported in part by National Natural Science Foundation of China (Nos. 62101136 and 62306075), Natural Science Foundation of Shanghai (No. 21ZR1403600), Shanghai Municipal Science and Technology Major Project (No. 2018SHZDZX01) and ZJLab, and Shanghai Center
for Brain Science and Brain-inspired Technology. \emph{(Corresponding authors: Yiming Lei; Hongming Shan)}}
\thanks{T. Chen, C. Wang, Z. Chen, and H. Shan are with the Institute of Science and Technology for Brain-inspired Intelligence, Fudan University, Shanghai 200433, China (e-mails: chent21@m.fudan.edu.cn; chenhuiwang21@m.fudan.edu.cn; zhihaochen21@m.fudan.edu.cn; hmshan@fudan.edu.cn)}
\thanks{Y. Lei is with the Shanghai Key Lab of Intelligent Information Processing, School of Computer Science, Fudan University, Shanghai 200433, China (e-mail: ymlei@fudan.edu.cn)}
}
\maketitle
\begin{abstract}
Medical image segmentation has been significantly advanced with the rapid development of deep learning (DL) techniques. Existing DL-based segmentation models are typically discriminative; \ie, they aim to learn a mapping from the input image to segmentation masks. However, these discriminative methods neglect the underlying data distribution and intrinsic class characteristics, suffering from unstable feature space. In this work, we propose to complement discriminative segmentation methods with the knowledge of underlying data distribution from generative models.
To that end, we propose a novel \underline{h}ybr\underline{i}d \underline{diff}usion framework for medical image segmentation, termed \methodname, which can synergize the strengths of existing discriminative segmentation models and new generative diffusion models. \methodname comprises two key components: discriminative segmentor and diffusion refiner. First, we utilize any conventional trained segmentation models as discriminative segmentor, which can provide a segmentation mask prior for diffusion refiner. Second, we propose a novel binary Bernoulli diffusion model (\refiner) as the diffusion refiner, which can effectively, efficiently, and interactively refine the segmentation mask by modeling the underlying data distribution. Third, we train the segmentor and \refiner in an alternate-collaborative manner to mutually boost each other. Extensive experimental results on abdomen organ, brain tumor, polyps, and retinal vessels segmentation datasets, covering four widely-used modalities, demonstrate the superior performance of \methodname over existing medical segmentation algorithms, including the state-of-the-art transformer- and diffusion-based ones. 
In addition, \methodname excels at segmenting small objects and generalizing to new datasets. Source codes are made available at \url{https://github.com/takimailto/HiDiff}.

\end{abstract}
\section{Introduction}
Medical image segmentation is to transform medical image data into meaningful, spatially structured information such as organs and tumors, which has been significantly advanced with the rapidly developed deep learning (DL) techniques~\cite{isensee2021nnu}. These DL-based segmentation methods have shown effectiveness in delineating organs/tumors and reducing labor costs.

Currently, existing DL-based segmentation methods, including convolutional neural networks-based~\cite{zhou2019unet++, huang2020unet,gu2020net}, and visual transformer-based variants~\cite{cao2022swin,chen2021transunet,dong2021polyp,huang2023missformer}, typically employ cross-entropy or Dice loss to learn a mapping function from the input medical image to the segmentation mask. Such a paradigm is often referred to as discriminative approaches that directly learn the classification probability of image pixels. Despite the prevalence, these approaches solely focus on learning the decision boundary between classes in the pixel feature space~\cite{liu2016large} and do not capture the underlying data distribution~\cite{bernardo2007generative}, failing to capture the intrinsic class characteristics. Furthermore, they learned an unstable feature space that results in a rapid performance drop when moving away from the decision boundaries~\cite{ardizzone2020training}, making it challenging to handle ambiguous boundaries and subtle objects.

In contrast, generative-based approaches~\cite{liang2022gmmseg,wolleb2022diffusion,austin2021structured} first model the joint probability of input data and segmentation mask, subsequently leverage the learned joint probability to evaluate the conditional distribution of segmentation masks given the input image, and finally output the mask prediction. A consensus emerging from numerous theoretical and empirical studies~\cite{efron1975efficiency,ng2001discriminative} suggests that generative-based methods possess the potential to mitigate the limitations associated with their discriminative counterparts because of the direct modeling of underlying data distribution.  However, it is worth noting that modern generative models also face challenges, including unstable training~\cite{goodfellow2020generative} and slow inference~\cite{ho2020denoising}. These challenges have prompted the exploration of integrated methods that combine discriminative and generative segmentation methods as a means to alleviate these issues~\cite{liang2022gmmseg,guo2022accelerating,wu2023medsegdiff, chen2023ascon}.

Recently, diffusion probabilistic model~(DPM)~\cite{sohl2015deep,ho2020denoising,song2020denoising} has shown impressive results in various image generation tasks~\cite{dorjsembe2022three, khader2022medical,gao2023corediff}, with powerful capabilities of modeling (un)conditional data distributions, which also facilitates the exploration of the field of medical image segmentation. On the one hand, one can directly apply the diffusion model to cast image segmentation as a generative task~\cite{amit2021segdiff,wu2022medsegdiff,wolleb2022diffusion}. On the other hand, some methods explore integrating the strengths of existing discriminative methods and generative methods to solve segmentation tasks. For example, PD-DDPM~\cite{guo2022accelerating} and MedSegDiff-V2~\cite{wu2023medsegdiff} try to tame the power of DPM with existing discriminative methods to improve the segmentation performance. However, most DPM-based segmentation methods rely on Gaussian noise as the diffusion kernel, neglecting the inherent \emph{discrete} nature of the segmentation task. In addition, the iterative denoising procedure of the DPM also makes the diffusion process time-consuming.

\begin{figure}[t]
	\centering
	\includegraphics[width=1 \linewidth]{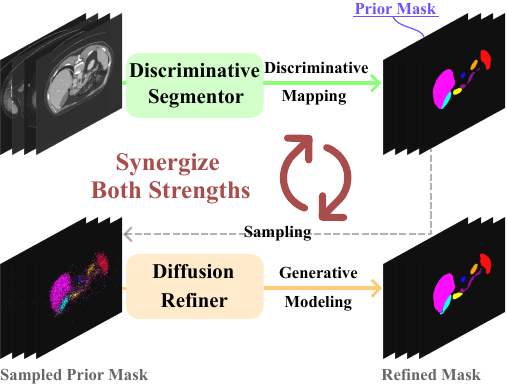}
 \caption{Conceptual illustration of our \methodname. We train our \methodname in an alternate-collaborative manner to synergize the strengths of existing discriminative segmentation and generative diffusion models.
}
\label{fig:concept} 
\end{figure}
In this work, we explore how to effectively, efficiently, and interactively synergize the strengths of existing discriminative segmentation models and generative DPM. To that end, we propose a novel \textbf{H}ybr\textbf{i}d \textbf{Diff}usion framework for medical image segmentation, termed \methodname, as shown in \figref{fig:concept}. The proposed \methodname involves two key components: discriminative segmentor and diffusion refiner. Our idea is to integrate the discriminative capacity of the existing segmentor with generative capacity of the diffusion refiner to improve medical image segmentation: the discriminative segmentor provides diffusion models with segmentation mask prior, and the diffusion refiner then  refines the segmentation mask effectively, efficiently, and interactively. We use existing medical image segmentation models as discriminative segmentor and propose a novel binary Bernoulli diffusion model (\refiner) as diffusion refiner. 
The novelty of \refiner lies in three-fold: (i) Bernoulli-based diffusion kernel enhances the diffusion models in modeling the discrete targets of the segmentation task, (ii) binarized diffusion refiner significantly improves efficiency for inference with negligible computational costs, and (iii) cross transformer enables interactive exchange between the diffusion generative feature and the discriminative feature. In addition, we train the discriminative segmentor and diffusion refiner of \methodname in an alternate-collaborative manner to mutually boost each other. 

The contributions of this work are summarized as follows.
\begin{enumerate}
    \item We propose a novel hybrid diffusion framework (\methodname) for medical image segmentation, which can synergize the strengths of existing discriminative segmentation models and the generative diffusion models. 
    \item We propose a novel binary Bernoulli diffusion model (\refiner) as the diffusion refiner, which can effectively, efficiently, and interactively refine the segmentation mask by modeling the underlying data distribution.
    \item We introduce an alternate-collaborative training strategy to train discriminative segmentor and diffusion refiner, which can improve each other during training.
    \item Extensive experimental results on abdomen organ~(Synapse), brain tumor~(BraTS-2021), polyps~(Kvasir-SEG and CVC-ClinicDB), and retinal vessels~(DRIVE and CHASE\_DB1) segmentation datasets demonstrate that \methodname achieves  superior performance over existing medical segmentation algorithms and excels at segmenting small objects and generalizing to new datasets. We highlight that \methodname is a principled framework that is fully compatible with existing discriminative segmentors.
\end{enumerate}

We note that a preliminary version of this work, Bernoulli Diffusion or \methodbase, was published in the International Conference on Medical Image Computing and Computer-Assisted Intervention~2023~\cite{chen2023berdiff}. In this paper, we further extend \methodbase~\cite{chen2023berdiff} with the following major improvements. First, we extend \methodbase~\cite{chen2023berdiff} to be a novel hybrid diffusion framework for medical image segmentation, leveraging both discriminative segmentation models and generative diffusion models, with a binarized diffusion refiner for improving both computational efficiency and usability. Second, \methodname leverages the proposed alternate-collaborative training strategy to facilitate the simultaneous training of the discriminative segmentor and diffusion refiner, leading to mutual performance improvements during training. Third, \methodname is a versatile framework that can seamlessly integrate with existing discriminative segmentation models, making it adaptable for a wide range of medical image segmentation tasks. Fourth, we extend \methodbase to be further compatible with multi-target segmentation tasks.

The remainder of this paper is organized as follows. We first detail the proposed methodology in \secref{sec:method}. We then elaborate on the experimental setup and results in \secref{sec:results}, followed by the discussion on related work and limitation in \secref{sec:discussion}. Finally, \secref{sec:conclusion} presents a concluding summary.

\begin{figure*}[t]
	\centering
	\includegraphics[width=1 \linewidth]{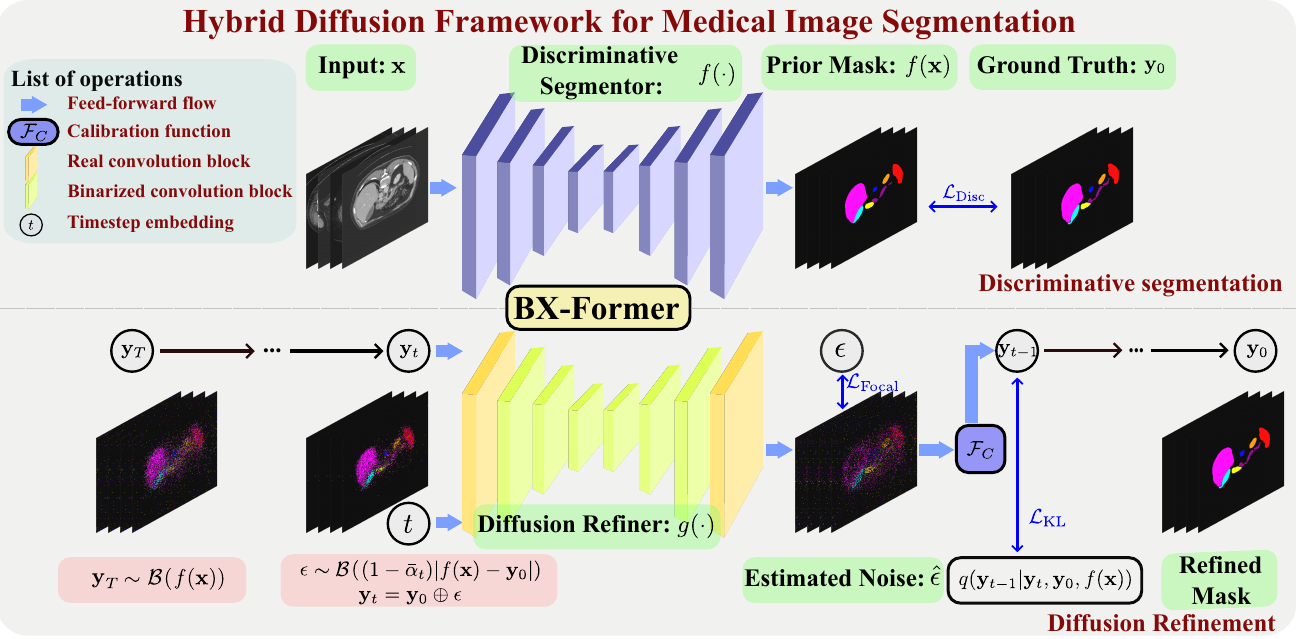}
	\caption{Overview of the proposed \methodname for medical image segmentation. Our \methodname involves two key components: discriminative segmentor and diffusion refiner, where the discriminative segmentor provides a segmentation mask prior for diffusion models while the diffusion refiner effectively, efficiently, and interactively refines the segmentation mask. Furthermore, we binarize our diffusion refiner and introduce a binary cross transformer to interactively exchange the discriminative and diffusion generative features, effectively refining the segmentation mask with negligible resources.}
\label{fig:overview} 
\end{figure*}
\section{Methodology}
\label{sec:method}
To effectively, efficiently, and interactively synergize the strengths of existing discriminative segmentors and the proposed \refiner, we propose a novel hybrid diffusion framework for medical image segmentation, as shown in \figref{fig:overview}. In this section, we first elaborate on existing discriminative segmentors and the proposed diffusion refiner in Secs.~\ref{sec:existing} and~\ref{sec:framework}, respectively. Finally, we detail our hybrid diffusion framework with an alternate-collaborative training strategy in \secref{sec:alternate}.

\subsection{Existing Discriminative Segmentors}
\label{sec:existing}

Let $\img{x} \in \mathbb{R}^{H{\times}W}$ denote the input medical image with a spatial resolution of $H{\times}W$, and $\img{y}_{0} \in \{0,1\}^{H{\times}W{\times}C}$ denote the ground-truth mask, where $C$ represents the number of classes. Given an image-mask pair $(\img{x}, \img{y}_{0})$, existing discriminative segmentation methods typically employ a deep neural network, $f(\cdot)$,  to predict the probability of segmentation mask, $f(\img{x}) \in \mathbb{R}^{H{\times}W{\times}C}$, with each element in a range of $(0, 1)$.
Most of them are trained end-to-end by minimizing cross-entropy loss or Dice loss.

Although such discriminative implementations usually perform satisfactorily, they still fail to capture the underlying data distribution and intrinsic class characteristics, resulting in an unstable feature space that makes it challenging to handle ambiguous boundaries and subtle objects. To address these limitations, we propose a hybrid diffusion framework to synergize the strengths of existing discriminative segmentation models and the proposed \refiner. For pre-training the discriminative segmentor used in our \methodname, we use the combination of cross-entropy loss $\loss{CE}$ and Dice loss $\loss{Dice}$ as the discriminative objective function:
\begin{align}
\loss{Disc} &= \loss{CE} + \lambda_\mathrm{Dice} \loss{Dice},
    \label{eqa:loss_disc}
\end{align}
where  $\lambda_\mathrm{Dice}$ is to adjust the weight of the Dice loss.

\subsection{Binary Bernoulli Diffusion Refiner}
\label{sec:framework}
Here, we propose a novel binary Bernoulli diffusion model (\refiner) as the diffusion refiner to effectively, efficiently, and interactively refine the prior mask generated by any discriminative segmentors through modeling the underlying data distribution. The novelty of \refiner lies in three-fold: (i) Bernoulli-based diffusion kernel to enhance the discrete capacity of diffusion models in modeling segmentation task, (ii) binarized diffusion refiner to significantly improve efficiency with negligible computational costs and, (iii) cross transformer to interactively exchange the diffusion generative feature with the discriminative feature for enhancement. In the following, we detail these three novelties.

\subsubsection{Bernoulli-based Diffusion Model for Effective Refinement}
\label{sec:berdiff}
We adopt a variant of U-Net~\cite{ronneberger2015u} as the diffusion refiner, $g(\cdot)$, to iteratively refine the prior mask $f(\img{x})$ generated by any discriminative segmentors $f(\cdot)$. Regarding the role of the prior mask, we employ it not only as involved in the noise addition during the diffusion forward process but also as the starting sampling point of the diffusion reverse process. The whole diffusion process of \refiner can be represented as: 
\begin{align}
\img{y}_T\xrightleftharpoons[]{} \cdots\xrightleftharpoons[]{}\img{y}_t \xrightleftharpoons[q(\img{y}_t\mid\img{y}_{t-1},f(\img{x}))]{p_\theta(\img{y}_{t-1}\mid\img{y}_t,f(\img{x}))}\img{y}_{t-1}\xrightleftharpoons[]{} \cdots\xrightleftharpoons[]{} \img{y}_0,
\end{align}
where $\mat{y}_{1},\ldots,\mat{y}_{T}$ are latent variables with the same size of the mask $\mat{y}_{0}$, $q(\cdot)$ and $p_{\theta}(\cdot)$ represent the diffusion forward and reverse processes, respectively.

Most existing diffusion-based segmentation methods rely on Gaussian noise as the diffusion kernel and neglect the inherent discrete nature of the segmentation task. Therefore, we propose a Bernoulli-based approach, detailed as follows, to address this limitation.

\noindent\textbf{Diffusion forward process.}\quad
\label{sec:bernoulli_forward_process}
In the diffusion forward process, our diffusion refiner gradually adds more Bernoulli noise using a cosine noise schedule $\beta_{1},\ldots, \beta_{T}$~\cite{nichol2021improved}. The Bernoulli forward process $q(\mat{y}_{1:T}|\mat{y}_{0}, f(\img{x}))$ is given by:
\begin{align}
\label{eqa:forward_1t}
    q\left(\img{y}_{1: T}|\img{y}_{0}, f(\img{x})\right) &:= \prod\nolimits_{t=1}^{T}q\left(\img{y}_{t} |\img{y}_{t-1}, f(\img{x})\right), \\ 
    \label{eqa:forward_tt-1}
    q\left(\img{y}_{t}|\img{y}_{t-1}, f(\img{x})\right) &:= \mathcal{B}((1-\beta_{t}) \img{y}_{t-1}+\beta_{t}f(\img{x})),
\end{align}
where $\mathcal{B}((1-\beta_{t}) \img{y}_{t-1}+\beta_{t}f(\img{x}))$ denotes the Bernoulli distribution, whose probability mass function is defined as:
\begin{equation}
\begin{cases}
q\left(\img{y}_{t}=1|\img{y}_{t-1}, f(\img{x})\right) = (1-\beta_{t}) \img{y}_{t-1}+\beta_{t}f(\img{x}),\\
    q\left(\img{y}_{t}=0|\img{y}_{t-1}, f(\img{x})\right) = 1-(1\!-\!\beta_{t}) \img{y}_{t-1}\!-\!\beta_{t}f(\img{x}).
\end{cases}
\end{equation}
Using the notation $\alpha_{t}=1-\beta_{t}$ and $\bar{\alpha}_{t}= {\prod_{\tau=1}^{t}} {\alpha}_{\tau}$, we can sample $\img{y}_{t}$ with an arbitrary time step $t$ in a closed form: 
\begin{align}
	q\left(\img{y}_{t}|\img{y}_{0},f(\img{x})\right) = \mathcal{B}(\bar{\alpha}_{t} \img{y}_{0}+(1-\bar{\alpha}_{t})f(\img{x})),
 	\label{eqa:forward_arbitrary}
\end{align}
where the mean parameter of this Bernoulli distribution can be viewed as an interpolation between the prior mask $f(\img{x})$ and the ground-truth mask $\img{y}_{0}$  with the time step increasing. We can further use the Bernoulli-sampled noise $\img{\epsilon} \sim \mathcal{B}((1-\bar{\alpha}_{t})\left|f(\img{x})-\img{y}_{0}\right|)$ conditioned by the prior mask to reparameterize $\img{y}_{t}$ in \equref{eqa:forward_arbitrary} as $\img{y}_{0}\oplus\img{\epsilon}$, where $|\cdot|$ denotes the absolute value operation, and $\oplus$ denotes the logical operation of ``exclusive or (XOR)''. 

The Bernoulli posterior can be represented as:
\begin{align}
	q(\img{y}_{t-1}\!\mid\!\img{y}_{t}, \img{y}_{0}, f(\img{x}))\!=\!\mathcal{B}(\phi_{\text{post}}\left(\img{y}_{t}, \img{y}_{0}, f(\img{x})\right)),\label{eqa:posterior}
\end{align}
where $\phi_{\text{post}}(\vct{y}_{t}, \vct{y}_{0},f(\img{x})) = \|\{\alpha_{t}\conc{1-\img{y}_{t}}{\img{y}_{t}}+(1-\alpha_{t})|1-\img{y}_{t}-f(\img{x})|\} \odot
    \{\bar{\alpha}_{t-1}\conc{1-\img{y}_{0}}{ \img{y}_{0}}+(1-\bar{\alpha}_{t-1})\conc{1-f(\img{x})}{ f(\img{x})}\}\|_1$~\cite{hoogeboom2021argmax}.
Here, $\odot$ denotes the element-wise product, $\conc{\cdot}{\cdot}$ the matrix concatenation along the channel dimension, and $\|\cdot\|_1$ the $\ell_1$ normalization along the channel dimension. We note that the matrix dimensions do not match in the first addition operation; we broadcast the second matrix to match the dimension of the first matrix.

\noindent\textbf{Diffusion reverse process.}\quad 
The diffusion reverse process $p_{\theta}(\img{y}_{0:T} |  f(\img{x}))$ can also be viewed as a Markov chain that starts from the prior mask $\img{y}_{T}$---sampled from a Bernoulli distribution parameterized by a pre-trained segmentaion model $f(\img{x})\in \mathbb{R}^{H{\times}W{\times}C}$ or $\img{y}_{T}\!\sim\!\mathcal{B}(f(\img{x}))$---and progresses through intermediate latent variables constrained by the prior mask $f(\img{x})$ to learn underlying data distribution:
\begin{align}
    &p_{\theta}(\img{y}_{0: T}|f(\img{x})):=
    p(\img{y}_{T}|f(\img{x}))
     \prod\nolimits_{t=1}^{T} p_{\theta}(\img{y}_{t-1}|\img{y}_{t}, f(\img{x})),\\
    &p_{\theta}(\img{y}_{t-1}|\img{y}_{t}, f(\img{x})):=\mathcal{B}(\hat{\img{\mu}}(\img{y}_{t}, t, f(\img{x}))).
    \label{eqa:predicted_posterior}
\end{align}

Specifically, our diffusion refiner $g(\cdot)$ estimates the Bernoulli noise $\hat{\img{\epsilon}}(\img{y}_{t}, t, f(\img{x}))$ under the  $t$-th time step to reparameterize $\hat{\img{\mu}}(\img{y}_{t}, t, f(\img{x}))$ via a calibration function $\mathcal{F}_{C}$ as follows:
\begin{align}
    \hat{\img{\mu}}(\img{y}_{t}, t, f(\img{x})) &=
    \mathcal{F}_{C}(\img{y}_{t},\hat{\img{\epsilon}}(\img{y}_{t}, t, f(\img{x})))\notag\\
    &=
    \phi_{\text{post}}(\img{y}_{t}, |\img{y}_{t}-\hat{\img{\epsilon}}(\img{y}_{t}, t, f(\img{x}))|,f(\img{x})).
\end{align}
$\mathcal{F}_{C}$ aims to calibrate the latent variable $\img{y}_t$ to a less noisy latent variable $\img{y}_{t-1}$ in two steps: (i) estimating the mask $\img{y}_0$ by computing the absolute deviation between $\img{y}_t$ and the estimated noise $\hat{\epsilon}$; and (ii) estimating the distribution of $\img{y}_{t-1}$ by calculating the Bernoulli posterior, $q(\img{y}_{t-1}\!\mid\!\img{y}_{t},\img{y}_0, f(\img{x}))$, using \equref{eqa:posterior}.

\noindent\textbf{Diffusion objective function.}\quad
Based on the variational upper bound on the negative log-likelihood in previous DPM~\cite{austin2021structured}, given an image-mask pair $(\img{x}, \img{y}_{0})$ and $t$-th latent variable $\img{y}_{t}$, we adopt Kullback-Leibler (KL) divergence and focal loss~\cite{lin2017focal} to optimize our diffusion refiner as follows:
\begin{align}
     \loss{KL} &= \mathrm{KL}[q(\img{y}_{t-1} | \img{y}_{t}, \img{y}_{0}, f(\img{x})) || p_{\theta}(\img{y}_{t-1} | \img{y}_{t}, f(\mat{x}))],
\label{eqa:loss_KL}\\
    \loss{Focal} &= -\frac{1}{HW} {\sum\nolimits_{i,j}^{H,W}} 
\Big[{\epsilon}_{i,j}(1-\hat{\epsilon}_{i,j})^{\gamma}\log_{}\hat{\epsilon}_{i,j}\notag\\
&\quad\quad\quad\quad\quad\quad\quad+ (1-\epsilon_{i,j})\hat{\epsilon}_{i,j}^{\gamma}\log_{}{(1-\hat{\epsilon}_{i,j})}\Big],
\label{eqa:loss_focal}
\end{align}
where $\img{\epsilon}$ and $\hat{\img{\epsilon}}$ represent the corresponding ground-truth and estimated Bernoulli noise, respectively. $\gamma$ is used to balance the relative loss for well-classified and misclassified pixels. When $\gamma=0$, $\loss{Focal}$ reduces to binary cross-entropy loss $\loss{BCE}$ used in  BerDiff~\cite{chen2023berdiff}.
Finally, the diffusion objective function is defined as:
\begin{align}
\loss{Diff}=\loss{KL} + \lamda{Focal}\loss{Focal},
\label{eqa:loss_diff}
\end{align}
where $\lamda{Focal}$ is to adjust the weight of the focal loss.

\subsubsection{Binarization Module for Efficient Refinement}
\label{sec:binary}
To alleviate the computational burdens of the iterative diffusion process, following~\cite{le2023binaryvit}, we propose to binarize the diffusion refiner to be lightweight with customized time-dependent binarization (TB) and activation (TA) modules, making efficient refinement with negligible resources. Of note, the superscript $b$ and $r$ are used to denote binary and real-valued, respectively.

\noindent\textbf{Binarized calculation.}\quad 
For the binarized calculation, we first binarize the real-valued input tensors $\vct{U}^r$ and weights $\vct{W}^r$ as $\vct{U}^b$ and $\vct{W}^b$, respectively, the concrete implementation is the same as ~\cite{rastegari2016xnor}. Then, the computationally heavy floating-point matrix multiplications between the real-valued input tensors $\vct{U}^r$ and weights $\vct{W}^r$ can be replaced by lightweight bitwise $\operatorname{XNOR}$ and $\operatorname{popcount}$ operations~\cite{bulat2019xnor} between binary $\vct{U}^b$ and $\vct{W}^b$, defined as follows:
\begin{align}
	\vct{U}^{b} * \vct{W}^{b}=\operatorname{popcount}\left(\operatorname{XNOR}\left(\vct{U}^{b}, \vct{W}^{b}\right)\right).
\end{align}

\noindent\textbf{Time-dependent binarized modules.}\quad
\label{sec:ta_and_tb} To effectively adapt to the iterative nature of the time-step conditioned DPM, inspired by the adaptive instance normalization~\cite{huang2017arbitrary}, we design the TB module to binarize the input tensors and the TA module to activate the input tensors dynamically.

The TB module is implemented with the channel-wise time-dependent binary thresholds $\alpha_{i}$:
\begin{align}
u_{i}^{b}=\operatorname{sign}\left(u_{i}^{r}\right)=
\begin{cases}
+1, & \text{ if } u_{i}^{r}>\alpha_{i} \\
-1, & \text{ if } u_{i}^{r} \leq \alpha_{i}
\end{cases},
\end{align}
where $u_{i}^{b}$ and $u_{i}^{r}$ are the binary and real-valued representations of the same input tensors' element on the $i$-th channel, respectively. In addition, $\alpha_{i}$ is generated by a lightweight fully connected layer, which takes the time step $t$ as input.

Analogously, the TA module is implemented as:
\begin{align}
\operatorname{TA}\left(u_{i}\right)=
\begin{cases}
u_{i}-\gamma_{i}+\zeta_{i}, & \text { if } u_{i}>\gamma_{i} \\
\beta_{i}\left(u_{i}-\gamma_{i}\right)+\zeta_{i}, & \text { if } u_{i} \leq \gamma_{i}
\end{cases},
\end{align}
where $\gamma_{i}$ and $\zeta_{i}$ are the learnable shift parameters on the $i$-th channel, linearly transformed from the time step $t$. $\beta_{i}$ is the learnable scaling coefficient.

 \begin{figure}[t]
	\centering
	\includegraphics[width=1 \linewidth]{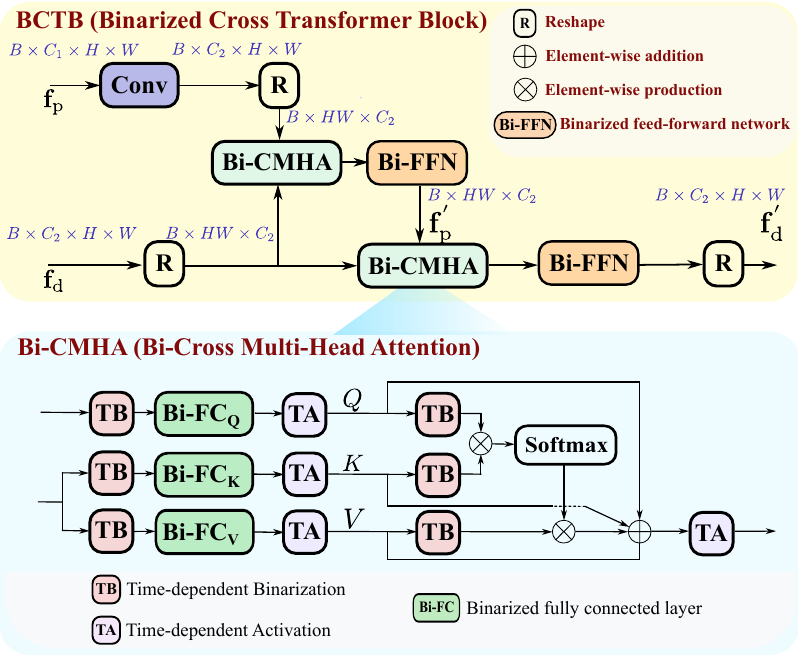}
 	\caption{Illustration of the proposed binarized cross transformer block and its constituent binarized cross multi-head attention modules.}
 	\label{fig:cross-attetion} 
\end{figure}

\subsubsection{Cross Transformer for Interactive Enhancement}
To interactively exchange the diffusion generative features with the discriminative features for enhancement, we propose a novel cross transformer, termed X-Former. The X-Former consists of two cross transformer blocks~(CTB). The first block encodes the feature with generative knowledge, extracted from the bottleneck of the U-shape diffusion refiner, into the middle position of the discriminative segmentor, \ie~$\vct{f}_\mathrm{d},\,\vct{f}_\mathrm{p}\rightarrow\vct{f}_\mathrm{p}^{\prime}$, while the second block operates in the opposite encoding direction, \ie~$\vct{f}_\mathrm{p}^{\prime}, \vct{f}_\mathrm{d} \rightarrow \vct{f}_\mathrm{d}^{\prime}$, injecting the features with the discriminative knowledge into the diffusion refiner. This bi-directional injection will achieve a stronger representation to generate a better-refined mask. 

Note that our X-Former can be binarized using the customized TA and TB modules to convert to its binarized counterpart, binarized X-Former or BX-Former, which consists of two binarized cross transformer blocks~(BCTB), as shown in \figref{fig:cross-attetion}. This binarization can be used to alleviate the computational burdens of the transformer block.

\begin{algorithm}[t]
\label{alg:training}
\small
	\SetAlgoLined
 \textbf{Input:} image-mask pairs $\{\Set{X}, \Set{Y}\}$, discriminative segmentor $f(\cdot)$, corresponding prior mask $f(\img{x})$, diffusion refiner $g(\cdot)$\\
 \textbf{Output:} optimized $f(\cdot)$ and $g(\cdot)$\\
	\For{each mini-batch}{
 \textbf{Sample} $(\img{x}, \vct{y}_{0})\!\sim\!\{\Set{X}, \Set{Y}\}$ $\&$ $t\!\sim\!\{1,\dots,T\}$\\
	
 \textbf{Sample} $\img{\epsilon}\!\sim\!\mathcal{B}((1-\bar{\alpha}_{t})\left|f(\img{x})-\img{y}_{0}\right|)$\\
 \textbf{Calculate}
 $\vct{y}_{t} = \vct{y}_{0} \oplus \vct{\epsilon}$\\
 	\tcp{Estimate the noise by the diffusion refiner $g(\cdot)$}
	\textbf{Estimate} $ \hat{\img{\epsilon}}(\img{y}_{t},t,f(\img{x}))$\\
	\tcp{Calculate the true and the estimated Bernoulli posterior}
	\textbf{Calculate} $q(\img{y}_{t-1}\!\mid\!\img{y}_{t}, \img{y}_{0}, f(\img{x}))\leftarrow$ \equref{eqa:posterior}\\
 \textbf{Calculate} $p_{\theta}(\img{y}_{t-1}\!\mid\!\img{y}_{t}, f(\img{x}))\leftarrow$ \equref{eqa:predicted_posterior}\\
\tcp{Optimizing diffusion refiner}
    \textbf{Calculate} $\loss{Diff}\leftarrow$ \equref{eqa:loss_diff}\\
	\textbf{Update} $g(\cdot)\leftarrow\nabla\loss{Diff}$ \textbf{while freezing} $f(\cdot)$ \\
    \tcp{Optimizing discriminative segmentor}
    \textbf{Calculate} $\loss{Hybrid}\leftarrow$ \equref{eqa:loss_prior}\\
    \textbf{Update} $f(\cdot)\leftarrow\nabla\loss{Hybrid}$ \textbf{while freezing} $g(\cdot)$\\}
	\caption{Training for \methodname}
\end{algorithm}

\subsection{Hybrid Diffusion Framework}
\label{sec:alternate}
To synergize the strengths of existing discriminative segmentation models and the proposed \refiner, we propose a hybrid diffusion framework for integrating the discriminative capacity of the existing segmentor with the generative capacity of the diffusion refiner for improved medical image segmentation, as shown in \figref{fig:overview}.  During discriminative segmentation, the discriminative segmentor provides a segmentation mask prior while during diffusion refinement, the \refiner acts as a diffusion refiner to effectively, efficiently, and interactively refine the segmentation mask. In the following, we detail the training and inference procedures of our \methodname, with the concrete algorithms in Algs.~\ref{alg:training} and~\ref{alg:sampling}.

During training, we optimize the diffusion refiner and the discriminative segmentor in an alternate-collaborative manner. 
Concretely, when optimizing the diffusion refiner, we freeze the discriminative segmentor and use the diffusion objective function in~\equref{eqa:loss_diff}. When optimizing the discriminative segmentor, we freeze the diffusion refiner and use the discriminative and diffusion objective functions collaboratively, as shown in:
\begin{align}\loss{Hybrid}=\loss{Disc}+\lamda{Diff}\loss{Diff},
    \label{eqa:loss_prior}
\end{align}
where $\lamda{Diff}$ is a trade-off hyperparameter.

During inference, the discriminative segmentor is first utilized to generate the prior mask $f(\img{x})$, then our \methodname samples the initial latent variable $\img{y}_T$ from the prior mask, followed by the iterative refinement to get a better mask. Note that our \methodname is also capable of other DPM accelerating strategies, such as Denoising Diffusion Implicit Model~(DDIM)~\cite{song2020denoising} as shown in Alg.~\ref{alg:sampling}. We follow~\cite{song2020denoising} to derive DDIM's sampling strategy for our \methodname and set the hyperparameter $\sigma_{t}$ to be $\frac{1-\bar{\alpha}_{t-1}}{1-\bar{\alpha}_{t}}$ to make the reverse process less random.

\begin{algorithm}[t]
	\label{alg:sampling}
	\small
	\SetAlgoLined
  \mbox{\textbf{Input:} image $\vct{x}$, discriminative segmentor~$f(\cdot)$,}\\
 \, diffusion refiner $g(\cdot)$\\
  \textbf{Output:} predicted mask $\vct{y}_{0}$ corresponding to $\vct{x}$\\
    \tcp{Get the prior mask by $f(\cdot)$}
    \textbf{Calculate} $f(\img{x})$\\
	\tcp{Sample the initial latent variable}
	$\vct{y}_{T} \sim \mathcal{B}(f(\img{x}))$\\
 \tcp{Refine the prior mask iteratively}
	\For{$t = T$ to $1$}{
	\tcp{Estimate the Bernoulli noise by $g(\cdot)$}
	\textbf{Calculate} $ \hat{\epsilon}(\img{y}_{t},t,f(\img{x}))$\\
	\textbf{When using DDPM's sampling strategy}:\\
  $\vct{y}_{t-1}\!\sim\!\mathcal{B}(\hat{\vct{\mu}})$ and $\hat{\img{\mu}}=\mathcal{F}_{C}(\img{y}_{t},\hat{\img{\epsilon}})$\\
	\textbf{When using DDIM's sampling strategy}:\\
	$\vct{y}_{t-1}\!\sim\! \mathcal{B}(\sigma_{t}\vct{y}_{t}+(\bar{\alpha}_{t-1}\!-\!\sigma_{t}\bar{\alpha}_{t})|\vct{y}_{t}\!-\!\hat{\vct{\epsilon}}(\vct{y}_{t}, t, \img{x})|+((1-\bar{\alpha}_{t-1})-(1-\bar{\alpha}_{t})\sigma_{t})f(\img{x}))$}
	\caption{Inference for \methodname}
\end{algorithm}

\section{Experiment}
\label{sec:results}
In this section, we first elaborate on the experimental setup in \secref{sec:setup}. Then, we compare our \methodname with other state-of-the-art (SOTA) methods on four benchmark tasks in \secref{sec:sota}, followed by generalizability assessment in \secref{sec:cross} and an analysis of performance on small objects of interest in \secref{sec:small}. To conclude, we conduct ablation studies to demonstrate further the superiority of the Bernoulli diffusion kernel, the compatibility of the proposed \methodname, and the effectiveness of the introduced constituent including the diffusion refinement process, the alternate-collaborative training strategy, the focal loss, the X-Former, and the binarization strategy in \secref{sec:ablation}.
\subsection{Experimental Setup}
\label{sec:setup}
\begin{figure*}[t]
	\centering
	\includegraphics[width=1 \linewidth]{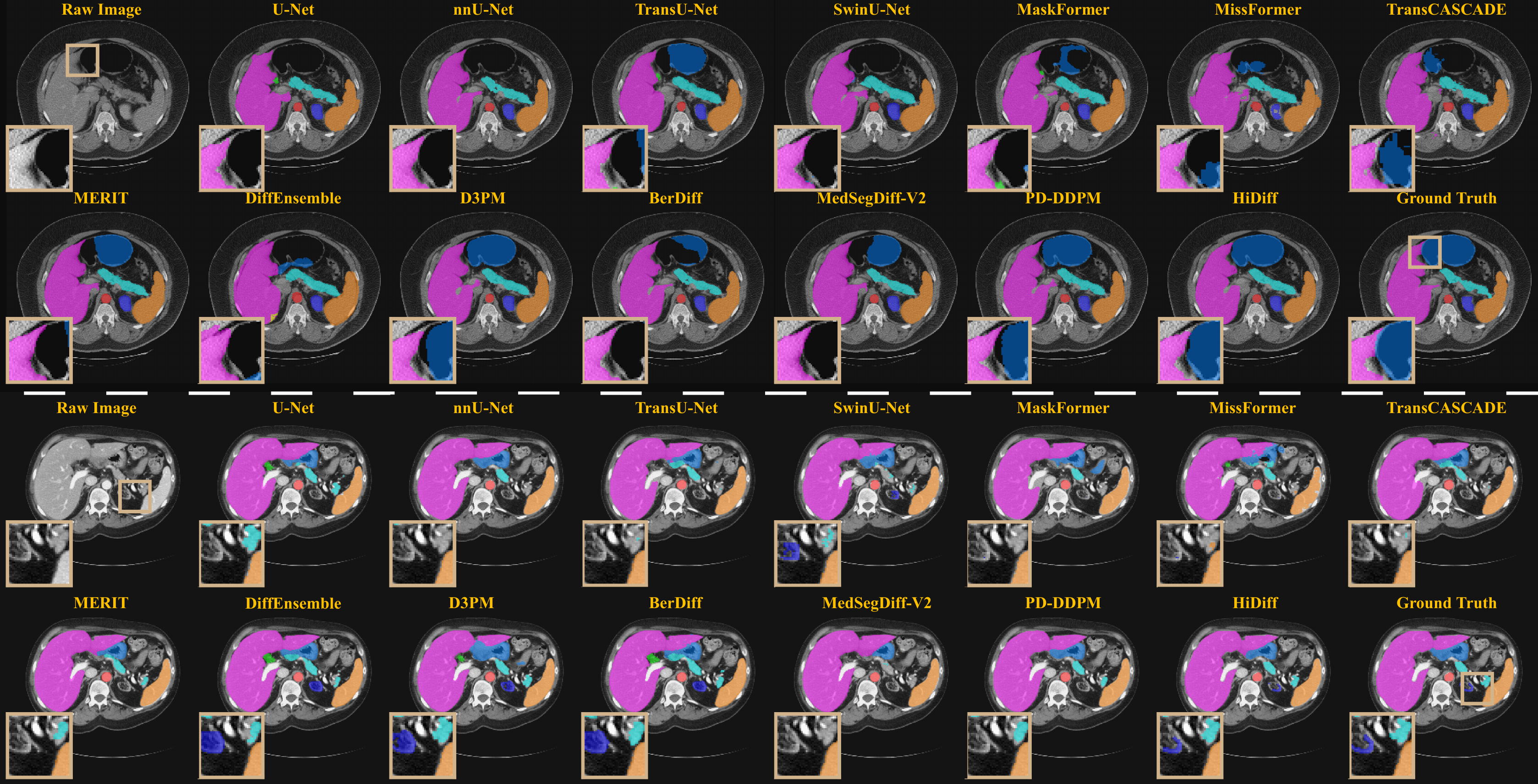}
 	\caption{Qualitative results of different segmentation methods for two cases, 0038 and 0008, from the Synapse testing set.}
  \label{fig:vis_synapse} 
  \end{figure*}
\begin{table*}[t]
	\centering
	\caption{Quantitative results on the Synapse testing set. The HD95 is represented as HD95(NaN Ratio). \textbf{I}, \textbf{II}, and \textbf{III} represent discriminative, generative diffusion, and integrated methods, respectively.}
	\label{tab:comparison_synapse}
	\begin{tabular*}{1\linewidth}{@{\extracolsep{\fill}}lccccccccccc}
	\shline
         {\multirow{2}{*}{\textbf{Methods}}}&
        &\multicolumn{2}{c}{\textbf{Average}}
        &\multicolumn{8}{c}{\textbf{Individual Dice}$\uparrow$}\\
        &&\textbf{Dice$\uparrow$}&\textbf{HD95$\downarrow$}&\textbf{Aorta}&\textbf{GB}&\textbf{KL}&\textbf{KR}&\textbf{Liver}&\textbf{PC}&\textbf{SP}&\textbf{SM}\\
         \cline{1-2}\cline{3-4}\cline{5-12}
        U-Net~\cite{ronneberger2015u}&{\multirow{7}{*}{\textbf{\large I}}}&76.75&27.94(2.08\%)&87.69&59.96&79.96&74.12&94.69&56.11&88.49&72.95\\
        nnU-Net~\cite{isensee2021nnu}&&\suboptimal{84.71}&19.74(1.04\%)&\best{91.84}&\best{77.91}&83.30&78.94&\suboptimal{96.48}&71.87&\best{92.10}&85.26\\
TransU-Net~\cite{chen2021transunet}&&83.51&21.82(1.04\%)&88.90&70.56&\best{86.69}&83.87&95.21&67.34&90.61&84.88\\
SwinU-Net~\cite{cao2022swin}&&81.13&18.30(1.04\%)&87.19&71.12&86.11&82.26&94.56&58.85&90.92&78.05\\
MaskFormer~\cite{cheng2021per}&&81.57&22.01(1.04\%)&87.14&70.47&82.42&77.31&95.07&65.42&90.56&84.16\\
        MissFormer~\cite{huang2023missformer}&&
        75.99& 22.04(1.04\%)&82.37&64.49&82.51&74.27&93.92&48.55&89.28&72.54\\
        TransCASCADE~\cite{rahman2023medical}&&81.55&17.59(1.04\%)&86.06&73.60&83.59&81.61&95.30&63.00&89.13&80.09\\
MERIT~\cite{rahman2023multi}&&83.75&16.45(1.04\%)&88.01&\suboptimal{74.85}&86.02&83.25&95.27&66.70&\suboptimal{91.01}&84.93\\
\hline
        DiffEnsemble~\cite{wolleb2022diffusion}&{\multirow{2}{*}{\textbf{\large II}}}&74.22&30.11(1.04\%)&86.78&51.94&84.60&76.32&92.60&46.07&86.74&68.69\\
        D3PM~\cite{austin2021structured}&&79.80&24.22(1.04\%)&83.13&61.30&80.82&79.61&95.27&68.12&89.47&80.65\\
        BerDiff~\cite{chen2023berdiff}&&82.45&\suboptimal{16.16}(1.04\%)&90.11&68.32&86.13&83.66&95.94&\suboptimal{68.64}&88.10&78.69\\
        \hline
        MedSegDiff-V2~\cite{wu2023medsegdiff}&{\multirow{3}{*}{\textbf{\large III}}}&78.16&28.06(1.04\%)&87.35&68.80&82.30&79.21&93.93&56.31&87.87&69.55\\
        PD-DDPM~\cite{guo2022accelerating}&&84.02&17.75(1.04\%)&88.91&73.49&86.13&\suboptimal{84.18}&95.46&67.33&89.49&\suboptimal{87.10}\\
        \methodname (\textbf{ours})&&\best{84.94}&\best{13.42}(1.04\%)&\suboptimal{90.17}&74.58&\suboptimal{86.28}&\best{84.59}&\best{96.66}&\best{69.48}&90.64&\best{87.12}\\
        \shline
		\end{tabular*}
\end{table*}

\subsubsection{Datasets} We conduct our experiments on four benchmark tasks, \ie~abdomen organ~(Synapse), brain tumor~(BraTS-2021), polyps~(Kvasir-SEG, CVC-ClinicDB) and retinal vessels~(DRIVE, CHASE\_DB1) segmentation datasets, covering four widely-used modalities such as  computed tomography (CT), magnetic resonance imaging (MRI), endoscopic images, and retinal images.

The Synapse dataset\footnote{\url{https://www.synapse.org/\#!Synapse:syn3193805/wiki/217789}} comprises 30 contrast-enhanced abdominal CT scans with 3,779 axial images. Each CT scan has around 85 to 198 slices with a resolution of 512$\times$512. Following~\cite{chen2021transunet}, we split the dataset into 18 scans (2,211 axial slices) for training and the rest for testing. Meanwhile, we divide all scans as 2D slices along the axial direction and use them for segmenting eight abdominal organs: aorta, gallbladder~(GB), left kidney~(KL), right kidney~(KR), liver, pancreas~(PC), spleen~(SP), and stomach~(SM). 

The BraTS dataset~\cite{baid2021rsna} consists of four different MRI sequences~(T1, T2, FLAIR, T1CE) for each patient, all of them are concatenated as the input, and each slice has a resolution of 240$\times$240. We use 2D axial slices and discard the bottom 80 and top 26 slices~\cite{wolleb2022diffusion}. Our training set includes 55,174 2D slices from 1,126 patients, while the corresponding testing set comprises 3,991 2D slices from the other 125 patients. We use the BraTS for the segmentation of three kinds of tumors: Necrotic tumor core~(NT), peritumoral edema~(ED), and enhancing tumor~(ET).

The Kvasir-SEG~\cite{jha2020kvasir} and CVC-ClinicDB~\cite{bernal2015wm} datasets contain endoscopic images with corresponding polyps annotations. The Kvasir-SEG dataset comprises 1000 polyps images with varying resolutions, while the CVC-ClinicDB dataset consists of 612 polyps images with a fixed resolution. Following~\cite{dumitru2023using}, we split both datasets into training, validation, and testing subsets according to a ratio of 80\%:10\%:10\%.

The DRIVE~\cite{staal2004ridge} and CHASE\_DB1 datasets~\cite{carballal2018automatic} contain retinal images with corresponding retinal vessels annotations. The DRIVE dataset contains 40 retinal images with a resolution of 565$\times$584 pixels while the CHASE\_DB1 dataset contains 28 retina images of 999$\times$960 pixels resolution. Following~\cite{liu2022full}, we adhere to the same dataset splitting. For the DRIVE dataset which has two manual annotations, we adopt the first one as the ground-truth mask~\cite{rahman2024g}.

\begin{figure*}[t]
	\centering
	\includegraphics[width=1 \linewidth]{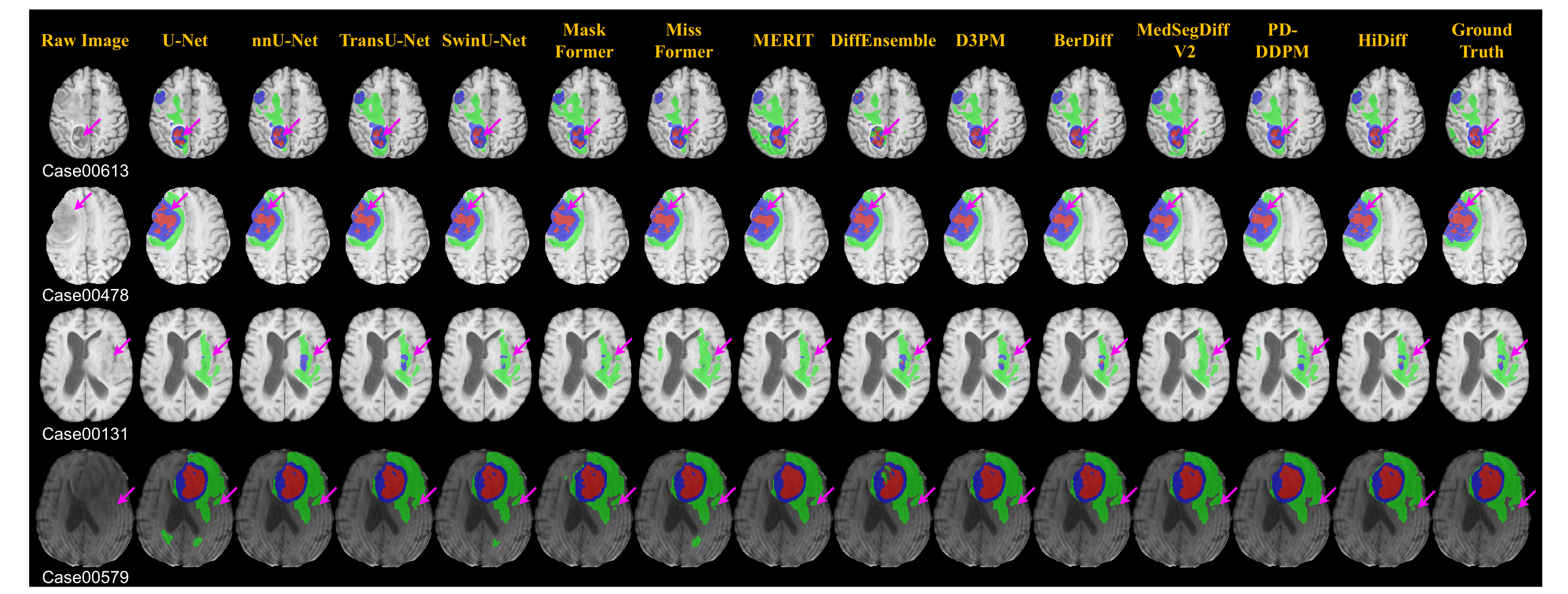}
 	\caption{Qualitative results of different segmentation methods for four cases from the BraTS testing set.}
 	\label{fig:vis_brats} 
\end{figure*}

\subsubsection{Implementation Details}
We implement all the methods with the PyTorch library~\cite{paszke2019pytorch} and train the networks on one NVIDIA V100 GPU. 
All the networks are trained using the AdamW~\cite{loshchilov2017decoupled} optimizer with a mini-batch size of 32. 
We set the hyperparameter $\gamma=2.0$ in \equref{eqa:loss_focal},  $\lambda_\mathrm{Dice}=1.0$ in \equref{eqa:loss_disc}, $\lambda_\mathrm{Focal}=1.0$ in \equref{eqa:loss_diff}, and $\lambda_\mathrm{Diff}=1.0$ in \equref{eqa:loss_prior}. The initial learning rate is set to $1\times10^{-4}$ for all datasets. We employ a cosine noise schedule~\cite{nichol2021improved} of $T=10$ timesteps for training our \methodname and $T=1,000$ timesteps for other competing diffusion models with  $
    \bar{\alpha}_{t}=\cos (\tfrac{t / T+s}{1+s} \cdot \tfrac{\pi}{2})^{2}/{\cos (\tfrac{s}{1+s} \cdot \tfrac{\pi}{2})^{2}}$, where $s=0.008$. For all integrated methods, we first pre-train the discriminative segmentor for 40,000 iterations with an initial learning rate of $1\times10^{-4}$ before proceeding with diffusion refiner. For a fair comparison, our \methodname and other diffusion-based models all use the respacing strategy of DDIM~\cite{song2020denoising} to accelerate the inference process with $T=10$ sampling steps. To reduce GPU memory usage during training, we train our \methodname on the Synapse dataset at 256$\times$256 resolution and test it at 512$\times$512 resolution~\cite{rahman2023multi,cao2022swin,rahman2023medical}. Simultaneously, for the BraTS dataset, we use 224$\times$224 resolution for both training and testing~\cite{wolleb2022diffusion,wu2023medsegdiff,chen2023berdiff}. For the Kvasir-SEG and CVC-ClinicDB datasets, the images are rescaled to 256$\times$256 resolution for both training and testing. For the DRIVE and CHASE\_DB1 datasets, we optimize the model by randomly cropping images to a resolution of 256$\times$256 during training, and evaluate the model by sliding a window of size 256$\times$256 over the original resolution (\ie~565$\times$584 for the DRIVE and 999$\times$960 for the CHASE\_DB1) during testing.

\subsubsection{Evaluation Metrics}
Following~\cite{chen2021transunet}, we adopt two widely used performance metrics for multi-object evaluation: Dice coefficient~(Dice)~\cite{taha2015metrics} and 95\% Hausdorff distance~(HD95)~\cite{huttenlocher1993comparing}. Dice calculates the overlap area between each sub-binary predicted and corresponding ground-truth mask. The HD95 represents the $95$th percentile of the Hausdorff distance, which measures the maximum distance between the predicted and ground truth mask. To address undefined HD95 when the label or prediction is $0^{H{\times}W}$, we only average over the defined case and further report the undefined number as NaN Ratio. For single-object evaluation, in addition to Dice and HD95, we further adopt IoU, recall, and accuracy based on previous papers~\cite{dumitru2023using,rahman2024g}. For all quantitative results, the best results are highlighted in {bold}. Note that all the experimental results are obtained through our own experiments, the difference in the results between ours and the original papers may be attributed to different experimental settings.

\subsection{Comparison to SOTA Methods}
\label{sec:sota}
The competing methods can be categorized into three groups: traditional discriminative methods, generative diffusion methods, and integrated methods. Concretely, discriminative methods include U-Net~\cite{ronneberger2015u}, nnU-Net~\cite{isensee2021nnu}, TransU-Net~\cite{chen2021transunet}, SwinU-Net~\cite{cao2022swin}, MaskFormer~\cite{cheng2021per}, MissFormer~\cite{huang2023missformer}, TransCASCADE~\cite{rahman2023medical}, and MERIT~\cite{rahman2023multi}. Generative diffusion methods include Gaussian-based DiffEnsemble~\cite{wolleb2022diffusion}, category-based D3PM~\cite{austin2021structured}, and Bernoulli-based BerDiff~\cite{chen2023berdiff}. Moreover, we compare our \methodname with recently published SOTA integrated methods, including MedSegDiff-V2~\cite{wu2023medsegdiff}, and PD-DDPM~\cite{guo2022accelerating}. Unless otherwise noted, we choose MERIT as the default discriminative segmentor for all the integrated methods and equipped \methodname with the binarized \refiner. For PD-DDPM, we set the starting sampling point at $300$ as suggested by their paper.

\begin{table}[t]
	\centering
	\caption{Quantitative results on the BraTS testing set. HD95 is represented as HD95(NaN Ratio). \textbf{I}, \textbf{II}, and \textbf{III} represent discriminative, generative diffusion, and integrated methods, respectively.}
	\label{tab:comparison_brats}
 \scriptsize
		\begin{tabular*}{1\linewidth}{@{\extracolsep{\fill}}lcccccc}
			\shline
			{\multirow{2}{*}{\textbf{Methods}}}&
			&\multicolumn{2}{c}{\textbf{Average}}
		      &\multicolumn{3}{c}{\textbf{Individual Dice}$\uparrow$}\\
&&\textbf{Dice$\uparrow$}&\textbf{HD95$\downarrow$}&\textbf{NT}&\textbf{ED}&\textbf{ET}\\
\hline
            U-Net~\cite{ronneberger2015u}&\multirow{6}{*}{\textbf{\large I}}&75.48&6.14(10.16\%)&69.60&77.87&78.98\\
        TransU-Net~\cite{chen2021transunet}&&83.50&3.65(5.97\%)&82.28&82.55&85.66\\
            SwinU-Net~\cite{cao2022swin}&&83.94&3.38(5.89\%)&82.28&83.57&85.99\\
            MaskFormer~\cite{cheng2021per}&&81.55&3.68(4.92\%)&80.24&81.34&83.05\\
    MissFormer~\cite{huang2023missformer}&&82.23&3.86(6.50\%)&80.70&82.32&83.67\\
            nnU-Net~\cite{isensee2021nnu}&&84.57&3.40(5.27\%)&\suboptimal{84.21}&83.39&86.12\\
            MERIT~\cite{rahman2023multi}&&83.02&3.70(5.70\%)&82.10&81.78&85.18\\
            \hline
    DiffEnsemble~\cite{wolleb2022diffusion}&\multirow{2}{*}{\textbf{\large II}}&71.00&6.50(13.66\%)&55.14&78.73&79.14\\
    D3PM~\cite{austin2021structured}&&84.88&\suboptimal{3.30}(4.87\%)&82.85&\suboptimal{84.68}&\suboptimal{87.10}\\
    BerDiff~\cite{chen2023berdiff}&&\suboptimal{85.42}&\best{3.03}(4.78\%)&84.36&84.61&87.29\\
            \hline
            MedSegDiff-V2~\cite{wu2023medsegdiff}&\multirow{3}{*}{\textbf{\large III}}&82.81&3.95(5.68\%)&82.61&83.84&77.95\\
            PD-DDPM~\cite{guo2022accelerating}&&83.51&3.63(5.62\%)&83.39&81.51&85.63\\
            {\methodname (\textbf{ours})}
            &&\best{85.80}&3.14(\best{3.60\%})&\best{84.23}&\best{85.30}&\best{87.78}\\
			\shline
		\end{tabular*}
\end{table}

\subsubsection{Results on the Synapse Dataset}
\tabref{tab:comparison_synapse} and \figref{fig:vis_synapse} present the quantitative and qualitative results on the Synapse dataset, respectively. We draw five key observations as follows. (i) For discriminative segmentor, the nnU-Net achieves the best performance because of the self-configuration and ensembling of five models. The transformer-based MERIT achieves the second-best performance as it leverages the power of multi-scale hierarchical visual transformers to better learn local and global relationships among pixels. (ii) The Gaussian-based generative diffusion methods, \ie~DiffEnsemble, cannot surpass most discriminative segmentor, even the vanilla U-Net baseline, which can be attributed to the deficiency of Gaussian diffusion kernel to handle the discrete nature of segmentation tasks. (iii) The discrete diffusion-based models, D3PM and BerDiff, outperform Gaussian-based models due to the introduced discrete category/Bernoulli diffusion kernel. (iv) For integrated methods, both PD-DDPM and our \methodname can synergize the strengths of existing discriminative segmentor and generative diffusion models while generating refined masks better than the initial discriminative segmentor. Another integrated method, MedSegDiff-V2, fails to surpass its discriminative counterpart, MERIT. This can be traced back to its initial sampling of pure Gaussian noise, hindering its ability to effectively utilize the segmentation mask prior. (v) Our \methodname outperforms all competing methods in terms of Dice and HD95, demonstrating the effectiveness of our \methodname.

The qualitative results in \figref{fig:vis_synapse} also highlight the effectiveness of our \methodname in handling complex anatomical structures, including the ambiguous boundary of large objects and subtle objects. Concretely, we observe that discriminative segmentation methods cannot cover the entire complex organ or even detect it at all, which could be attributed to the inherently unstable feature space learned by this discriminative training paradigm. In contrast, most generative ones have better coverage due to their capacity to model the underlying data distribution. However, they still face challenges in addressing ambiguous boundaries and subtle objects. By synergizing the strengths of both discriminative segmentor and generative diffusion model, the integrated methods, especially ours, can not only accurately delineate the ambiguous boundaries of large organs, \eg~the stomach in Case 0038, but are also good at detecting subtle organs, \eg~the left kidney in Case 0008. By further comparing our \methodname with its discriminative counterpart MERIT, we observe that our \methodname can indeed refine the segmentation mask to be visually more accurate.

\subsubsection{Results on the BraTS Dataset} 
\tabref{tab:comparison_brats} and \figref{fig:vis_brats} present the quantitative and qualitative results on the BraTS-2021 dataset, respectively, which also demonstrate the superior performance of our \methodname in discrete segmentation tasks, in line with the analysis on the Synapse dataset. Note that MedSegDiff-V2 performs better on the BraTS than on the Synapse due to the simpler segmentation relationship in the the BraTS dataset.

\figref{fig:vis_brats} shows that our \methodname performs well in the detection of tumors with subcategories; \eg~successfully identifying small tumor lesions, such as necrotic tumor cores in Cases 00613 and 00478, as well as peritumoral edema in Case 00131. In addition, our method accurately delineates the boundaries of larger tumor lesions, such as enhancing tumors in Case 00579. Taking Case 00579 as an example, many discriminative segmentation methods suffer from false positives in detecting enhancing tumors due to the limited modeling of the underlying data distribution. In contrast, both the generative diffusion and integrated methods, exhibit a notable absence of such false positives. Furthermore, our \methodname excels in precisely delineating boundaries and subtle objects.

For the following experiments, we only select MERIT, BerDiff, and PD-DDPM as the baselines because they are top performers across discriminative, generative, and integrated segmentation methods on the Synapse and BraTS-2021.

\subsubsection{Results on the Kvasir-SEG and CVC-ClinicDB Dataset}

\begin{figure}[!t]
\centering
\includegraphics[width=1\linewidth]{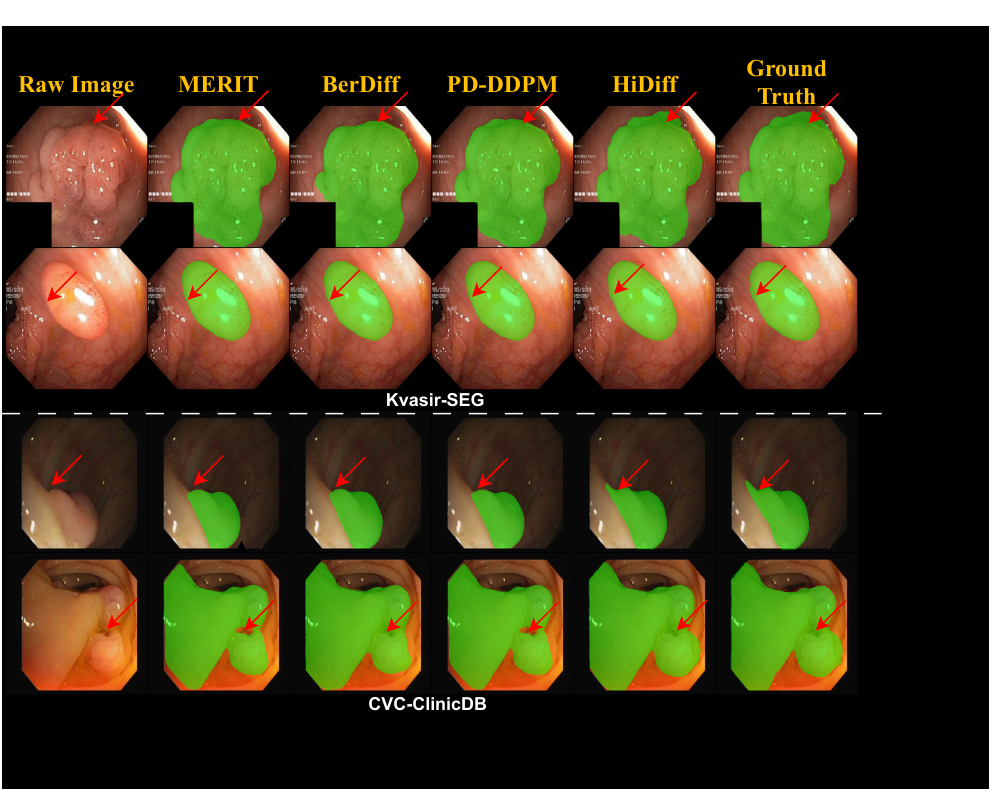}
\caption{Qualitative results of different segmentation methods for four cases from the Kvasir-SEG and CVC-ClinicDB testing sets.}
\label{fig:vis_Polyp}
\end{figure}

\tabref{tab:comparison_polyp} and \figref{fig:vis_Polyp} present the quantitative and qualitative results on the Kvasir-SEG and CVC-ClinicDB datasets, respectively, which demonstrate the superior performance of our \methodname for single-object segmentation tasks. \figref{fig:vis_Polyp} shows that our \methodname performs well in the outline of boundaries. As indicated by the red arrows, \methodname accurately delineates the boundaries (1st and 3rd rows), and also rationally fills cavities (4th row).

\begin{table*}[!htb]
	\centering
	\caption{Quantitative results on the Kvasir-SEG and CVC-ClinicDB testing sets. NaN ratios are omitted because they are all 0.}
	\label{tab:comparison_polyp}
 \scriptsize
		\begin{tabular*}{1\linewidth}{@{\extracolsep{\fill}}lcccccccccc}
			\shline
			{\multirow{2}{*}{\textbf{Methods}}}&\multicolumn{5}{c}{\textbf{Kvasir-SEG}}&\multicolumn{5}{c}{\textbf{CVC-ClinicDB}}\\
&\textbf{Dice$\uparrow$}&\textbf{HD95$\downarrow$}&\textbf{IoU$\uparrow$}&\textbf{Recall$\uparrow$}&\textbf{Accuracy$\uparrow$}&\textbf{Dice$\uparrow$}&\textbf{HD95$\downarrow$}&\textbf{IoU$\uparrow$}&\textbf{Recall$\uparrow$}&\textbf{Accuracy$\uparrow$}\\
  \cline{1-1}\cline{2-6}\cline{7-11}MERIT~\cite{rahman2023multi}&95.32&11.45&91.07&\suboptimal{95.12}&\suboptimal{98.50}&95.73&5.49&91.80&94.97&99.25\\
  BerDiff~\cite{chen2023berdiff}&94.52&12.43&89.61&93.41&98.26&\suboptimal{95.80}&6.77&\suboptimal{91.94}&\suboptimal{95.53}&\suboptimal{99.26}\\
  PD-DDPM~\cite{guo2022accelerating}&\suboptimal{95.33}&11.45&\suboptimal{91.08}&95.13&\suboptimal{98.50}&95.73&5.47&91.81&94.98&99.25\\
  \methodname (\textbf{ours})&\best{96.29}&\best{11.08}&\best{92.00}&\best{96.36}&\best{99.48}&\best{96.21}&\best{5.36}&\best{93.57}&\best{96.45}&\best{99.50}\\
\shline
\end{tabular*}
\end{table*}

\begin{table*}[!t]
	\centering
	\caption{Quantitative results on the DRIVE and CHASE\_DB1 testing sets. NaN ratios are omitted because they are all 0.}
	\label{tab:comparison_retinal}
 \resizebox{\textwidth}{!}{
 \scriptsize
	\begin{threeparttable}
		\begin{tabular*}{1\linewidth}{@{\extracolsep{\fill}}lcccccccccc}
			\shline
			{\multirow{2}{*}{\textbf{Methods}}}&\multicolumn{5}{c}{\textbf{DRIVE}}&\multicolumn{5}{c}{\textbf{CHASE\_DB1}}\\
&\textbf{Dice$\uparrow$}&\textbf{HD95$\downarrow$}&\textbf{IoU$\uparrow$}&\textbf{Recall$\uparrow$}&\textbf{Accuracy$\uparrow$}&\textbf{Dice$\uparrow$}&\textbf{HD95$\downarrow$}&\textbf{IoU$\uparrow$}&\textbf{Recall$\uparrow$}&\textbf{Accuracy$\uparrow$}\\
  \cline{1-1}\cline{2-6}\cline{7-11}
  MERIT~\cite{rahman2023multi}&80.24&8.88&67.00&80.21&96.54&80.27&14.50&67.04&83.21&97.42\\
  BerDiff~\cite{chen2023berdiff}&\suboptimal{80.51}&4.66&\suboptimal{67.37}&\suboptimal{81.30}&96.55&79.54&17.49&66.03&\suboptimal{87.48}&97.16\\
  PD-DDPM~\cite{guo2022accelerating}&80.24&8.21&67.00&80.21&\suboptimal{96.56}&\suboptimal{80.39}&20.52&\suboptimal{67.21}&83.21&\suboptimal{97.44}\\
  \methodname (\textbf{ours})&\best{81.72}&\best{4.33}&\best{69.09}&\best{82.31}&\best{96.78}&\best{82.13}&\best{13.18}&\best{69.68}&\best{90.20}&\best{97.53}\\
\shline
\end{tabular*}
\end{threeparttable}}
\end{table*}

\subsubsection{Results on the DRIVE and CHASE\_DB1}
\label{sec:retinal}
\begin{figure}[!t]
\centering

\includegraphics[width=1 \linewidth]{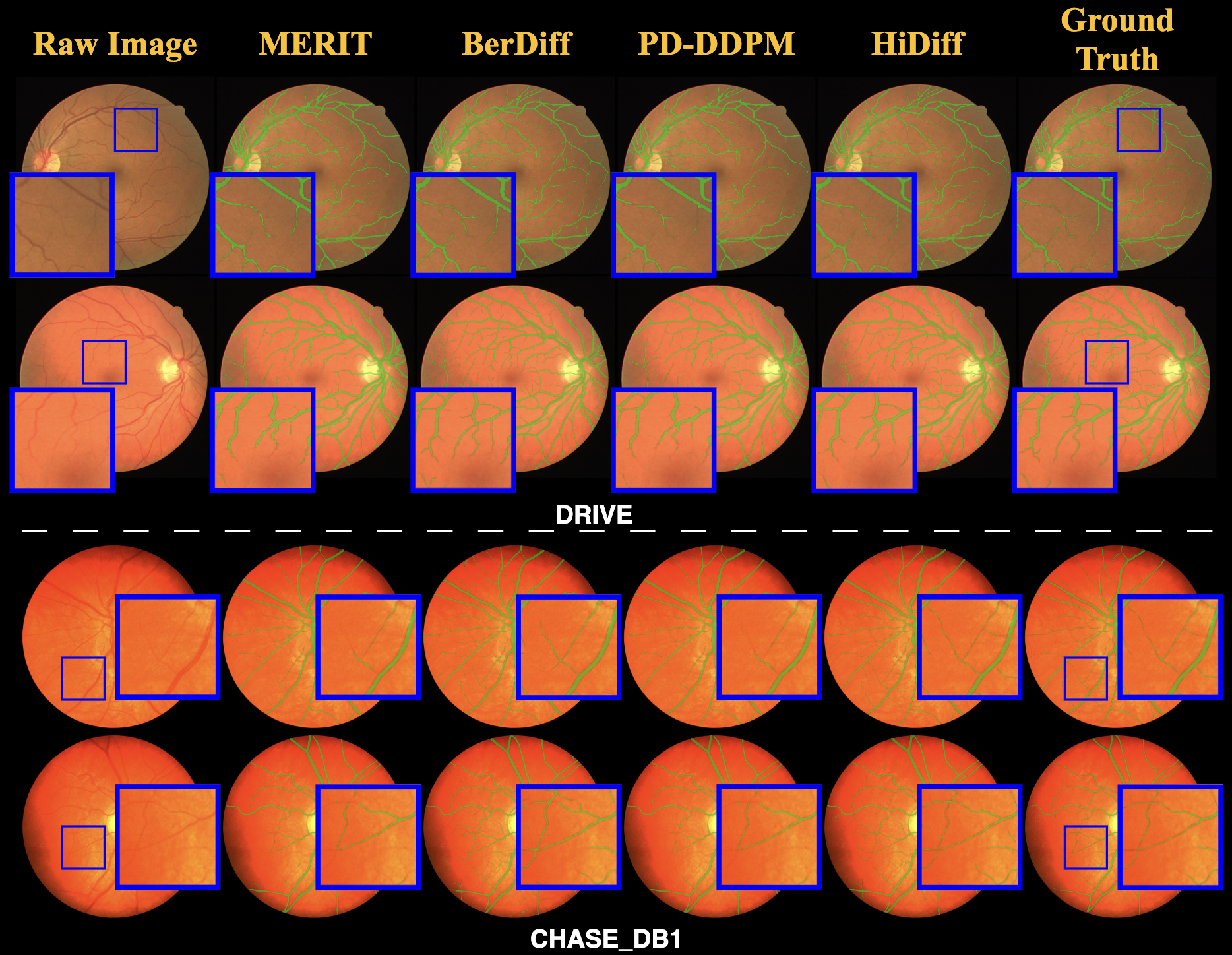}
\caption{Qualitative results of different segmentation 
methods for four cases from the Drive and CHASE\_DB1 testing sets.}
\label{fig:vis_Retinal}
\end{figure}

\tabref{tab:comparison_retinal} and \figref{fig:vis_Retinal} present the quantitative and qualitative results on the retinal vessels segmentation task, respectively. The results show that our \methodname achieves better performance and highlight the better capability of our \methodname to outline small structures, \ie~the thin and intricate blood vessels.

In summary, the quantitative and qualitative results on four datasets show that our \methodname consistently outperforms its discriminative counterpart, MERIT, and highlights the robustness and versatility of our \methodname, with great potential for challenging segmentation tasks.

\subsection{Cross-dataset Evaluation}
\label{sec:cross}
Here, we perform cross-dataset evaluations to further assess the generalizability of our \methodname. Specifically, we conduct cross-dataset evaluation on {(i)} \emph{abdomen CT organ segmentation task}, training on the Synapse while testing on the Medical Segmentation Decathlon (MSD)~\cite{simpson2019large}, and {(ii)} \emph{polyps segmentation task} using the Kvasir-SEG and CVC-ClinicDB datasets with two different directions, \ie~training on the Kvasir-SEG while testing on the ClinicDB and training on the ClinicDB while testing on the Kvasir-SEG.

\begin{table}[t]
	\centering
	\caption{Cross-dataset
 quantitative results of CT organ segmentation task. NaN ratios are omitted because they are all 0.}
	\label{tab:comparison_synapse_cross}
        \begin{threeparttable}
	\begin{tabular*}{1\linewidth}{@{\extracolsep{\fill}}lccccc}
	\shline
         {\multirow{2}{*}{\textbf{Methods}}}
        &\multicolumn{2}{c}{\textbf{Average}}
        &\multicolumn{3}{c}{\textbf{Individual Dice}$\uparrow$}\\
        &\textbf{Dice$\uparrow$}&\textbf{HD95$\downarrow$}&\textbf{Liver}&\textbf{PC}&\textbf{SP}\\
         \cline{1-1}\cline{2-3}\cline{4-6}
MERIT~\cite{rahman2023multi}&82.39&9.88&94.55&61.05&91.58\\
BerDiff~\cite{chen2023berdiff}&81.46&17.07&94.54&59.75&90.08\\
PD-DDPM~\cite{guo2022accelerating}&\suboptimal{82.60}&\suboptimal{9.84}&\suboptimal{94.77}&\suboptimal{61.25}&\suboptimal{91.78}\\
\methodname (\textbf{ours})&\best{84.19}&\best{8.58}&\best{95.62}&\best{63.69}&\best{93.27}\\
\shline
\end{tabular*}
\end{threeparttable}
\end{table}
\begin{table*}[!t]
	\centering
	\caption{
 Cross-dataset quantitative results of the polyps segmentation task using the Kvasir-SEG and CVC-ClinicDB datasets.  HD95 is represented as HD95(NaN Ratio).}
	\label{tab:comparison_polyp_cross}
	 \resizebox{\textwidth}{!}{
 \scriptsize
	\begin{threeparttable}
		\begin{tabular*}{1\linewidth}{@{\extracolsep{\fill}}lcccccccccc}
			\shline
			{\multirow{2}{*}{\textbf{Methods}}}&\multicolumn{5}{c}{\textbf{Kvasir-SEG$\rightarrow$ CVC-ClinicDB}}&\multicolumn{5}{c}{\textbf{CVC-ClinicDB$\rightarrow$Kvasir-SEG}}\\
&\textbf{Dice$\uparrow$}&\textbf{HD95$\downarrow$}&\textbf{IoU$\uparrow$}&\textbf{Recall$\uparrow$}&\textbf{Accuracy$\uparrow$}&\textbf{Dice$\uparrow$}&\textbf{HD95$\downarrow$}&\textbf{IoU$\uparrow$}&\textbf{Recall$\uparrow$}&\textbf{Accuracy$\uparrow$}\\
\cline{1-1}\cline{2-6}\cline{7-11}
MERIT~\cite{rahman2023multi}&\suboptimal{82.63}&18.46(1.61\%)&\suboptimal{70.39}&\suboptimal{77.20}&\suboptimal{97.13}&87.46&19.63(0.00\%)&77.72&80.67&96.27\\
BerDiff~\cite{chen2023berdiff}&73.65&24.56(3.23\%)&58.29&76.07&95.19&77.79&27.66(2.00\%)&63.65&67.75&93.77\\
PD-DDPM~\cite{guo2022accelerating}&82.62&19.20(1.61\%)&\suboptimal{70.39}&77.19&\suboptimal{97.13}&\suboptimal{87.47}&18.22(0.00\%)&\suboptimal{77.73}&\suboptimal{80.68}&\suboptimal{96.28}\\
\methodname (\textbf{ours})&\best{83.81}&\best{16.38}(1.61\%)&\best{71.63}&\best{79.50}&\best{97.39}&\best{88.38}&\best{16.96}(0.00\%)&\best{78.80}&\best{81.77}&\best{96.76}\\
\shline
\end{tabular*}
\end{threeparttable}}
\end{table*}
\begin{table*}[!ht]
	\centering
	\caption{Quantitative results on a small organ subset of the Synapse testing set. The HD95 is represented as HD95(NaN Ratio).}
	\label{tab:comparison_small_synapse}
 \resizebox{\textwidth}{!}{
 \scriptsize
        \begin{threeparttable}
	\begin{tabular*}{1\linewidth}{@{\extracolsep{\fill}}lcccccccccc}
	\shline
         {\multirow{2}{*}{\textbf{Methods}}}
        &\multicolumn{2}{c}{\textbf{Average}}
        &\multicolumn{8}{c}{\textbf{Individual Dice}$\uparrow$}\\
        &\textbf{Dice$\uparrow$}&\textbf{HD95$\downarrow$}&\textbf{Aorta}&\textbf{GB}&\textbf{KL}&\textbf{KR}&\textbf{Liver}&\textbf{PC}&\textbf{SP}&\textbf{SM}\\
         \cline{1-1}\cline{2-3}\cline{4-11}
MERIT~\cite{rahman2023multi}&29.44&16.44(50.48\%)&79.20&44.15&29.82&30.12&7.36&23.92&17.25&3.73\\
BerDiff~\cite{chen2023berdiff}&\suboptimal{44.18}&28.96(30.85\%)&\best{87.92}&\suboptimal{56.23}&\best{51.52}&\suboptimal{40.00}&\suboptimal{35.68}&\suboptimal{29.83}&\best{36.49}&\suboptimal{15.78}\\
PD-DDPM~\cite{guo2022accelerating}&30.29&\best{16.28}(49.84\%)&80.69&44.56&30.45&31.76&7.71&25.48&17.38&4.25\\
\methodname (\textbf{ours})&\best{44.70}&20.73(\best{26.45\%})&\suboptimal{87.56}&\best{58.71}&\suboptimal{49.17}&\best{44.26}&\best{43.97}&\best{32.22}&\suboptimal{23.31}&\best{18.36}\\
\shline
\end{tabular*}
\end{threeparttable}}
\end{table*}

\begin{figure}[!t]
\centering
\includegraphics[width=1 \linewidth]{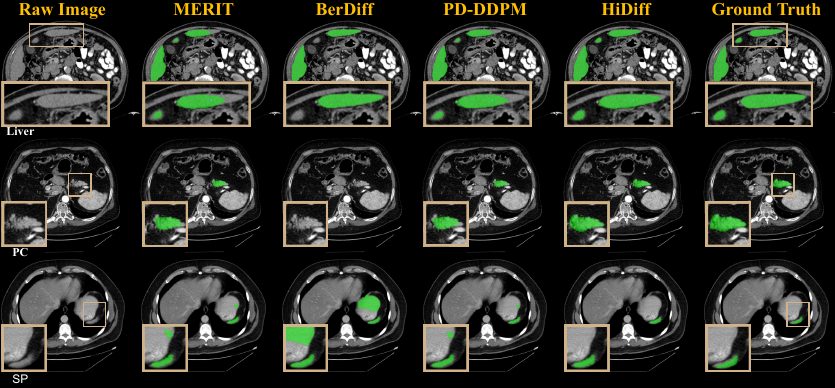}
\caption{Cross-dataset qualitative results of different segmentation methods for four cases from the MSD testing set.}
\label{fig:MSD_cross}
\end{figure}
\begin{figure}[!t]
\centering
\includegraphics[width=1 \linewidth]{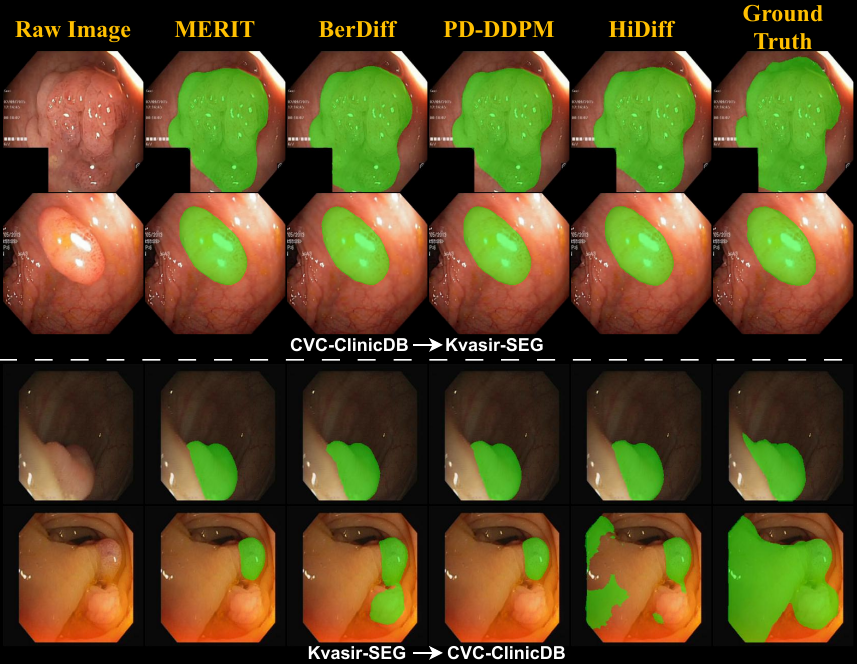}
\caption{Cross-dataset qualitative results of different methods on the polyps segmentation task.}
\label{fig:vis_Polyp_cross}
\end{figure}

\subsubsection{Abdomen CT Organ Segmentation} 
We train different models on the Synapse dataset and evaluate them on the MSD dataset with three organ segmentation tasks, including liver, PC, and SP; the same preprocessing pipelines is adopted for the Synapse and MSD datasets~\cite{chen2021transunet}. To facilitate testing, we randomly select 10 3D volumes for each sub-task. Table~\ref{tab:comparison_synapse_cross} and \figref{fig:MSD_cross} present the cross-dataset quantitative and qualitative results of the CT organ segmentation task, respectively. We find that our \methodname consistently outperforms its discriminative counterpart and achieves better generalizability. Our previous version, \methodbase,  exhibits subpar performance during cross-dataset evaluations, which may  be attributed to the fact that directly modeling the underlying data distribution without incorporating a discriminative prior may lead to overfitting and consequently yield poor results. However, in the case of our \methodname, leveraging the capabilities of discriminative segmentation models results in better generalization. Similar enhancements in generalization also manifest in other integrated methods, \eg~PD-DDPM. This phenomenon further validates that discriminative and generative models mutually benefit each other. Qualitative results in \figref{fig:MSD_cross} validate the superior performance of our \methodname on the external testing set of MSD. Specifically, taking 1st row as an example, our \methodname can not only accurately delineate large objects on the top, but also detect subtle objects in the middle.

\subsubsection{Polyps Segmentation} We use the Kvasir-SEG and CVC-ClinicDB datasets to evaluate the generalizability of the polyps segmentation task with two different directions. Table~\ref{tab:comparison_polyp_cross} and \figref{fig:vis_Polyp_cross} present corresponding quantitative and qualitative results, respectively. Despite the existing domain gap between these two datasets, our \methodname achieves better generalizability. We attribute this to the capability of our \methodname to model the underlying data distribution with the discriminative prior effectively, enhancing its generalizability across different datasets. From the qualitative perspective, taking 4th row in \figref{fig:vis_Polyp_cross} as an example, although it is difficult to segment for some cases with relatively large domain gap, our \methodname is still better than other competitors.

\begin{figure}[!t]
\centering
\includegraphics[width=1 \linewidth]{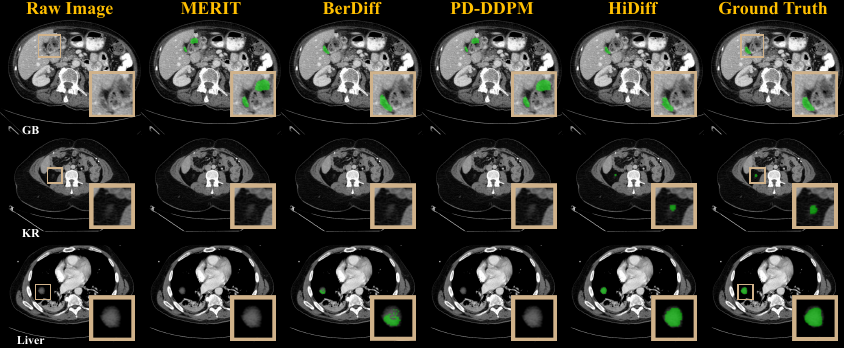}
\caption{Qualitative results of different segmentation methods for three cases from the small organ subset of the Synapse testing set.}
\label{fig:small_synapse}
\end{figure}

\subsection{Small Objects of Interest}
To further assess the capacity of our \methodname for small object segmentation, we perform evaluations on the small object subsets of the Synapse and BraTS testing sets.
\label{sec:small}
\subsubsection{Small Organs in the Synapse} 
We present quantitative results for a small organ subset of the Synapse testing set, with 2D organ pixel-wise size smaller than 500, as shown in \tabref{tab:comparison_small_synapse}. We can see that MERIT performs poorly for small organs and PD-DDPM also performs poorly for MERIT's bad initialization. While BerDiff is quite good in small objects segmentation with a surprisingly low NaN Ratio, our \methodname surpasses the baseline by utilizing the alternate-collaborative training strategy to maximize the synergy of discriminative segmentation and generative diffusion. The qualitative results presented in \figref{fig:small_synapse} also validate the superior small object detection capability of BerDiff and our \methodname, while our \methodname shows better boundary delineation.

\subsubsection{Small Tumors in the BraTS-2021}
\begin{figure}[!t]
\centering
\includegraphics[width=1 \linewidth]{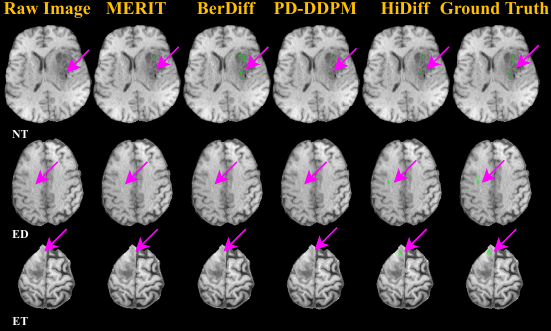}
\caption{Qualitative results of different segmentation methods for three cases from the small tumor subset of the BraTS testing set.}
\label{fig:small_brats}
\end{figure}

We test the performance of our \methodname on small tumor subset~(defined as $\text{pixel-wise tumor size}<100$), with quantitative results as shown in Table~\ref{tab:comparison_brats_small}. The results again validate the better performance on small objects segmentation. Qualitative results in \figref{fig:small_brats} highlight the superior small tumor detection of our \methodname as indicated by the arrows.
\begin{table}[!t]
	\centering
	\caption{Quantitative results on a small tumor subset of the BraTS testing set. The HD95 is represented as HD95(NaN Ratio).}
	\label{tab:comparison_brats_small}
	\begin{threeparttable}
		\begin{tabular*}{1\linewidth}{@{\extracolsep{\fill}}lccccc}
			\shline
			{\multirow{2}{*}{\textbf{Methods}}}
			&\multicolumn{2}{c}{\textbf{Average}}
		      &\multicolumn{3}{c}{\textbf{Individual Dice}$\uparrow$}\\
&\textbf{Dice$\uparrow$}&\textbf{HD95$\downarrow$}&\textbf{NT}&\textbf{ED}&\textbf{ET}\\
\hline
MERIT~\cite{rahman2023multi}&31.91&8.82(43.55\%)&33.75&29.64&32.34\\
BerDiff~\cite{chen2023berdiff}&\suboptimal{46.62}&8.81(27.29\%)&\suboptimal{55.02}&\suboptimal{39.52}&\suboptimal{45.31}\\
PD-DDPM~\cite{guo2022accelerating}&33.35&\best{8.58}(44.32\%)&36.64&30.27&33.14\\
{\methodname (\textbf{ours})}&\best{51.77}&8.76(\best{20.13\%})&\best{59.00}&\best{47.30}&\best{49.00}\\
			\shline
		\end{tabular*}
	\end{threeparttable}
\end{table}

With the experimental results for retinal vessels segmentation in \secref{sec:retinal}, we highlight the versatility of our \methodname in accurately segmenting small objects across diverse medical modalities, \eg~CT, MRI, and retinal images.

\subsection{Ablation Study}
\label{sec:ablation}
In this subsection, we first investigate the effects of diffusion kernels and compatibility with other discriminative segmentors. 
We then conduct a series of ablation experiments to validate the effectiveness of key components in our \methodname, including 
 the diffusion refinement, the alternate-collaborative training strategy, the focal loss, and the X-Former. 
Unless otherwise noted, all ablation experiments are conducted on the Synapse dataset.

\subsubsection{Effects of Diffusion Kernel (Gaussian v.s. Bernoulli)}

We initiate our ablation by establishing the superiority of the Bernoulli diffusion kernel over its Gaussian counterpart for segmentation tasks, which is based on experiments without the discriminative segmentors, \ie~the same as traditional BerDiff~\cite{chen2023berdiff}. Specifically, we compare the performance of Bernoulli diffusion~(BerDiff) against that of Gaussian diffusion~(DiffEnsemble) on the Synapse and BraTS datasets in Tables~\ref{tab:comparison_synapse} and~\ref{tab:comparison_brats}. We found that Bernoulli diffusion consistently outperforms Gaussian diffusion, highlighting the advantages of the discrete kernel for segmentation tasks. This finding serves as a foundation for the subsequent advancements.

\subsubsection{Compatibility of the \methodname}

\begin{figure}[t]
\centering

\includegraphics[width=1 \linewidth]{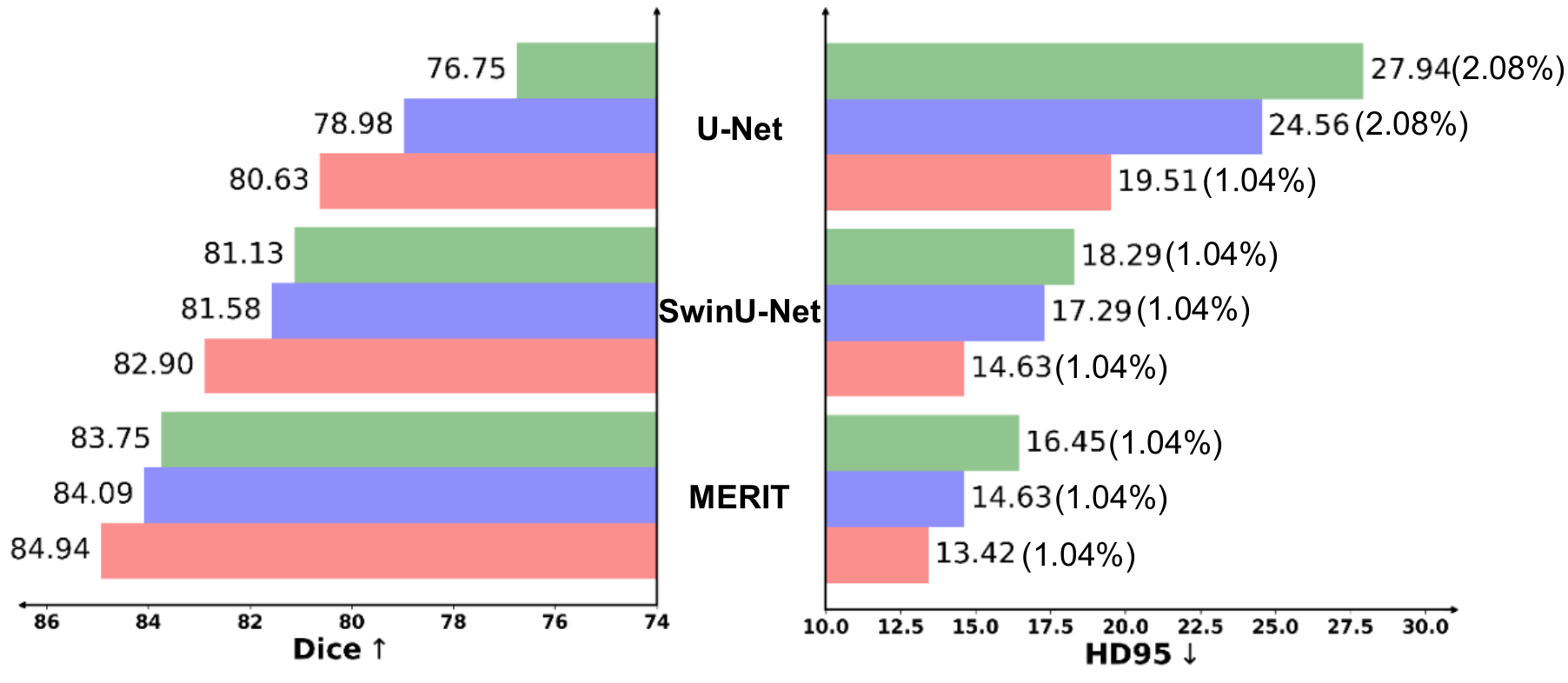}
\caption{Quantitative results of our \methodname with three different discriminative segmentors: U-Net, SwinU-Net, and MERIT. HD95 is represented as HD95(NaN Ratio).}
\label{fig:bar_v1}
\end{figure}

To validate the compatibility of our \methodname, we combine the proposed BBDM with three typical discriminative segmentors: U-Net, SwinU-Net, and MERIT. \figref{fig:bar_v1} presents the quantitative results on the Synapse dataset, showing that our \methodname not only consistently improves segmentation performance with diffusion refiner but also boosts the discriminative segmentor itself through the introduced alternate-collaborative training strategy. This suggests that our \methodname is a principled framework fully compatible with any SOTA discriminative segmentors.

\begin{table}[t]
	\centering
	\caption{Ablation results of diffusion refinement. NaN Ratios are omitted because they are all 1.04\%.}
	\label{tab:comparison_refinement}
	\begin{threeparttable}
		\begin{tabular*}{1\linewidth}{@{\extracolsep{\fill}}lcc}
			\shline
			{\textbf{Variants}}&\textbf{Dice$\uparrow$}&\textbf{HD95$\downarrow$}\\
			\cline{1-3}
MERIT~\cite{rahman2023multi} (baseline)&83.75&16.45\\
\quad + discriminative refinement &\suboptimal{83.82}&\suboptimal{16.38}\\
\quad + diffusion refinement (\textbf{ours})&\best{84.94}&\best{13.42}\\
\shline
		\end{tabular*}
	\end{threeparttable}
\end{table}

\subsubsection{Ablation on Diffusion Refinement}
We validate the effectiveness of the proposed diffusion refinement by replacing the proposed BBDM with a discriminative refinement network. Here, we use a binarized U-Net as a discriminative refiner while keeping other settings consistent, such as both utilizing MERIT as the discriminative segmentor. \tabref{tab:comparison_refinement} presents the results, which show that both the discriminative refinement and \methodname yield better performance than baseline MERIT. However, the improvement of the discriminative refinement falls far short of the one achieved by our diffusion refinement process. This highlights the effectiveness of our diffusion refinement, emphasizing that performance improvements are not achieved simply by increasing the number of parameters.

\begin{figure*}[t]
\centering
\includegraphics[width=1\linewidth]{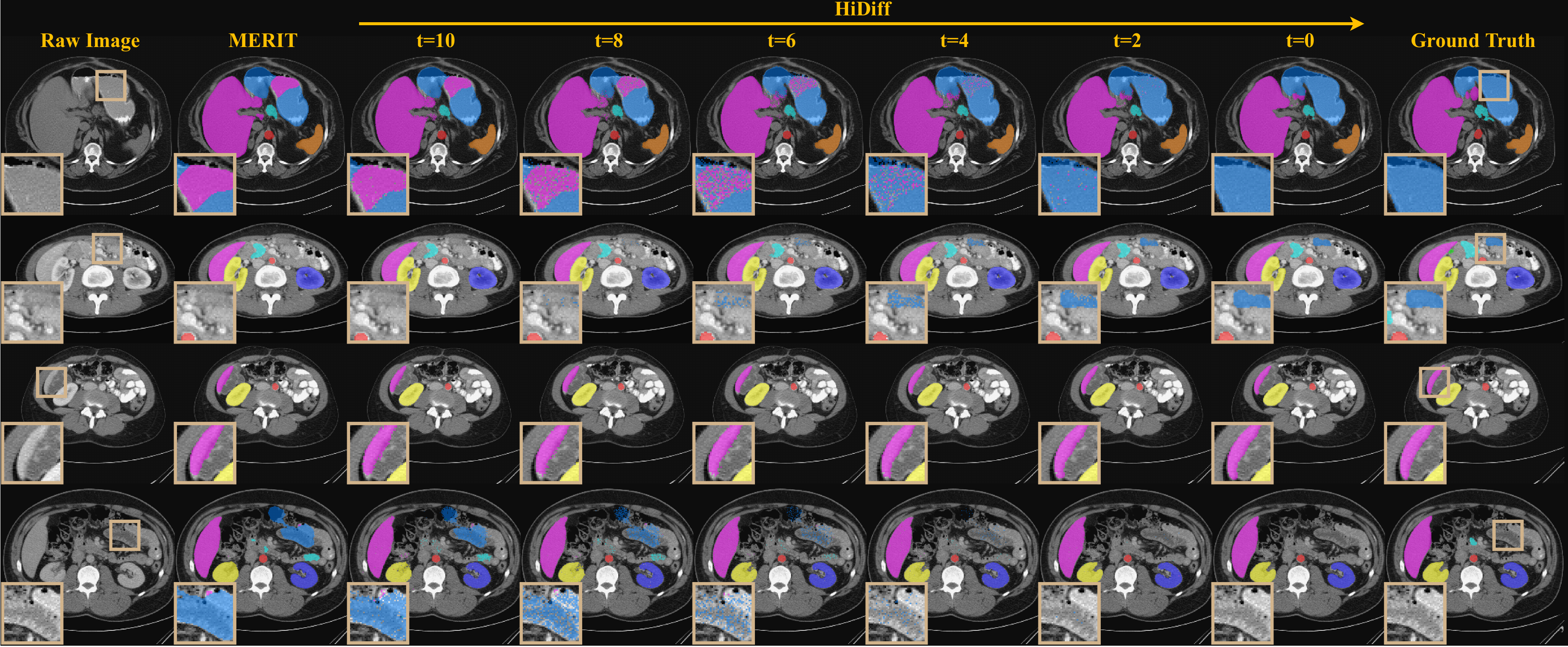}
\caption{Qualitative case study on the effectiveness of diffusion refinement process for four cases from the Synapse testing set (0025, 0004, 0002, and 0036).}
\label{fig:vis_ablation}
\end{figure*}

Furthermore, \figref{fig:vis_ablation} presents the qualitative case study on the effectiveness of diffusion refinement, demonstrating that our \methodname can refine incorrect segment regions iteratively, such as rectifying incorrect organ layouts in Case 0025, identifying previously undetected organs in Case 0004, improving boundary accuracy in Case 0002, and eliminating falsely detected organs in Case 0036. These observations suggest the significance of introducing the knowledge of the underlying data distribution from the diffusion refiner into the discriminative segmentor. Such introduction can empower the model to gain deeper modeling of intrinsic class characteristics and, consequently, achieve a more stable feature space.

\subsubsection{Ablation on Alternate-collaborative Training Strategy}
We proceed to validate the effectiveness of the alternate-collaborative training strategy by comparing V1 and V2 in \tabref{tab:ablation_loss}. Both variants employ the non-binarized \refiner without X-Former as the diffusion refiner and were trained with loss functions combining $\loss{KL}$ and $\loss{BCE}$, with or without the alternate-collaborative training strategy, respectively. The comparison results reveal a performance improvement when such a strategy is employed. We attribute it to the fact that such an alternate-collaborative training strategy prompts bi-directional knowledge distillation between the discriminative segmentor and the diffusion refiner.

\begin{table}[t]
	\centering
	\caption{Ablation results of training strategy, focal loss, X-Former, and binarization. V1, V2, V3, V4, and V5 represent 5 implementations. NaN Ratios are omitted because they are all 1.04\%. }
	\label{tab:ablation_loss}
		\begin{tabular*}{1\linewidth}{@{\extracolsep{\fill}}lcccccc}
			\shline\textbf{}&{\textbf{Loss}}&{\textbf{Alternate}}&\textbf{X-Former}&\textbf{binarized}
			&\textbf{Dice$\uparrow$}&\textbf{HD95$\downarrow$}\\
			\hline
V1&$\loss{BCE}$&\XSolidBrush&\XSolidBrush&\XSolidBrush&84.02&15.05\\
V2&$\loss{BCE}$&\CheckmarkBold&\XSolidBrush&\XSolidBrush&84.54&14.52\\
V3&$\loss{Focal}$&\CheckmarkBold&\XSolidBrush&\XSolidBrush&84.73&\suboptimal{13.40}\\
V4&$\loss{Focal}$&\CheckmarkBold&\CheckmarkBold&\XSolidBrush&\best{85.11}&\best{12.35}\\
V5&$\loss{Focal}$&\CheckmarkBold&\CheckmarkBold&\CheckmarkBold&\suboptimal{84.94}&13.42\\
\shline
		\end{tabular*}
\end{table}

\begin{table}[t]
	\centering
	\caption{FLOPS~(${\times}10^{10}$) and storage~(MB) of different methods. Note that for integrated methods, FLOPs are presented as FLOPs of discriminative segmentor + FLOPs of a single forward process of the diffusion refiner× diffusion steps, and storage is presented as storage of discriminative segmentor + storage of diffusion refiner.}
	\label{tab:ablation_binarize}
		\begin{tabular*}{1\linewidth}{@{\extracolsep{\fill}}lcll}
			\shline
			\textbf{Methods}
		&\textbf{binarized}
  &\textbf{FLOPs}$\downarrow$&\textbf{Storage$\downarrow$}\\
			\hline
      U-Net~\cite{ronneberger2015u}&\XSolidBrush&0.32&4.6\\
   nnU-Net~\cite{isensee2021nnu}&\XSolidBrush&11.52${\times}$5&268.7$\times$5\\
   
   TransU-Net~\cite{chen2021transunet}&\XSolidBrush&3.23&355.7\\
   SwinU-Net~\cite{cao2022swin}&\XSolidBrush&0.59&103.6\\
   MaskFormer~\cite{cheng2021per}&\XSolidBrush&6.55&349.5\\
   MissFormer~\cite{huang2023missformer}&\XSolidBrush&0.73&135.2\\
   TransCASCADE~\cite{rahman2023medical}&\XSolidBrush&3.23&355.7\\
   MERIT~\cite{rahman2023multi}&\XSolidBrush&6.66&563.4\\
   \hline
   DiffEnsemble~\cite{wolleb2022diffusion}&\XSolidBrush&26.5$\times$10&326.5\\
   D3PM~\cite{austin2021structured}&\XSolidBrush&26.4$\times$10&326.4\\
   BerDiff~\cite{chen2023berdiff}&\XSolidBrush&26.4$\times$10&326.4\\
   \hline
   PD-DDPM~\cite{guo2022accelerating}&\XSolidBrush&6.66$+$26.5$\times$10&563.4$+$326.5\\
   MedSegDiff-V2~\cite{wolleb2022diffusion}&\XSolidBrush&6.66$+$23.8$\times$10&563.4$+$433.5\\
   \methodname (\textbf{ours})& \XSolidBrush&6.66$+$8.93$\times$10&563.4$+$230.6\\
   \methodname (\textbf{ours})& \CheckmarkBold&6.66$+$0.41$\times$10{\best{(\textbf{22$\times$})}}&563.4$+\,\,\,$18.6\\
\shline
\end{tabular*}
\end{table}

\subsubsection{Ablation on Focal Loss}
Here, we provide a quantitative analysis involving different loss components to explore the impact of selecting the focal loss. The results are presented in V2 and V3 of \tabref{tab:ablation_loss}. These two variants employ the non-binarized \refiner without X-Former and are trained with loss functions that combine $\loss{KL}$ and $\loss{BCE}$, or $\loss{KL}$ and $\loss{Focal}$, respectively. We note that $\loss{Focal}$ reduces to $\loss{BCE}$ when $\gamma=0$ in Eq.~\eqref{eqa:loss_focal}. The comparison between V2 and V3 reveals a performance improvement when the focal loss was employed. We attribute it to the fact that our \methodname introduces Bernoulli noise based on the prior mask, resulting in smaller perturbations compared to traditional BerDiff, which further introduces a class-imbalance challenge during training.

\subsubsection{Ablation on  X-Former}
We then present the ablation results regarding X-Former, as summarized in V3 and V4 of \tabref{tab:ablation_loss}. Both are trained with a loss function that combines $\loss{KL}$ and $\loss{Focal}$ and employ the non-binarized \refiner with X-Former or without X-Former. The comparison reveals that the introduction of X-Former can contribute to an enhancement in segmentation results by the bi-directional injection between the discriminative and diffusion generative features.

\subsubsection{Ablation on Binarization}
Here, we discuss the performance and the computational cost of the binarization, and the corresponding results are presented in V4 and V5 of \tabref{tab:ablation_loss} and \tabref{tab:ablation_binarize}, respectively.
Regarding the performance, V4 and V5 of \tabref{tab:ablation_loss} are trained with loss functions that combine $\loss{KL}$ and $\loss{Focal}$ and employ the non-binarized or binarized \refiner with X-Former. We observe that the introduction of binarization exerts a negligible influence on the segmentation performance. We attribute this phenomenon to our \methodname's intrinsic advantage---a robust prior mask that provides a stable baseline, and the fact that the binary calculations align well with the Bernoulli diffusion process.

Regarding the computational cost, we further compare the floating point operations~(FLOPs) and memory consumption changes of binarization in the last four rows of \tabref{tab:ablation_binarize}. Note that for modern computers, 64 binary operations~(BOPs) are equivalent to 1 FLOP, and we follow this convention. We observe that our non-binarized \methodname already has fewer FLOPs, almost $3\times$ faster inference, compared to existing integrated methods, due to the lightweight choice of building modules. Furthermore, when combined with binarization, our \methodname further gets $22\times$ speed-up for the diffusion refinement process, and 10$\times$ speed-up for the entire inference process, effectively reducing the computational burdens. In summary, our \methodname achieves \emph{significant performance improvements} at a cost of a \emph{relatively modest increase in computational burden}.
\section{Discussion}
\label{sec:discussion}
\subsection{Difference from Bernoulli Diffusion~\cite{sohl2015deep}}
\label{related_work}
Bernoulli diffusion has already been introduced in~\cite{sohl2015deep} for deep unsupervised learning. The key differences between our \methodname and~\cite{sohl2015deep} are summarized as follows. (i) Our \methodname synergizes the strengths of existing discriminative segmentation models and new generative diffusion models while~\cite{sohl2015deep} solely focuses on the conventional generative capability. (ii) Our \methodname introduces a novel parameterization technique---namely the calibration function---for estimating the Bernoulli noise of $\img{y}_{t}$, which has been shown more effective than directly estimating $\img{y}_{0}$~\cite{sohl2015deep} in the preliminary version~\cite{chen2023berdiff}. (iii) We highlight that our \methodname is the first to apply Bernoulli diffusion to the medical image segmentation task while~\cite{sohl2015deep} primarily focuses on deep unsupervised learning for unconditional image generation.

\subsection{Limitation}
We acknowledge that our \methodname bears the computational burden of both the discriminative segmentor and the diffusion refiner, as shown in \tabref{tab:ablation_binarize}. \emph{From a speed perspective}, we can find that the computational burden for other integrated methods primarily resides in the diffusion refiner, whereas for our \methodname, the discriminative segmentor (MERIT) significantly contributes to the overall computational demand. \emph{For memory consumption}, although our \methodname's memory requirements are larger compared to other discriminative or generative segmentation methods, it is comparable to other integrated methods. The selection of a large discriminative segmentor contributes to this memory consumption. We emphasize that this choice was made to harness the robustness of MERIT as a backbone for our framework. However, we emphasize that our \methodname is a versatile framework and fully compatible with any SOTA discriminative segmentors and there is room for exploring smaller and more memory-efficient discriminative segmentors like~\cite{rahman2024g} in the future research.
\section{Conclusion}
\label{sec:conclusion}
This paper proposed a novel hybrid diffusion framework, \methodname, for medical image segmentation, which can synergize the strengths of existing discriminative segmentation models and new generative diffusion models, \ie~\refiner. The novelty of our \refiner lies in three-fold: (i) effective: Bernoulli-based diffusion kernel to enhance the diffusion models in modeling the discrete targets of the segmentation task, (ii) efficient: the binarized diffusion refiner to significantly improve efficiency for inference with negligible computational costs, and (iii) interactive: cross transformer to enable interactive exchange between the diffusion generative feature and the discriminative feature. We train \methodname in an alternate-collaborative manner, which can mutually boost the discriminative segmentor and the diffusion refiner during training. Extensive experimental results and detailed ablation studies validated the superior performance of \methodname and the effectiveness of key components in \methodname. We highlight that \methodname is a principled framework fully compatible with existing DL-based segmentation models.

\end{document}